\documentclass[lettersize,journal]{IEEEtran}
\usepackage{amsmath,amsfonts}
\usepackage{algorithmic}
\usepackage{algorithm}
\usepackage{array}
\usepackage[caption=false,font=normalsize,labelfont=sf,textfont=sf]{subfig}
\usepackage{textcomp}
\usepackage{stfloats}
\usepackage{url}
\usepackage{verbatim}
\usepackage{graphicx}
\usepackage{cite}
\usepackage{bm}
\usepackage{multirow}
\usepackage{xcolor}
\usepackage{makecell}
\usepackage{algorithm}
\usepackage{algorithmic}
\usepackage{subcaption}
\usepackage{adjustbox}

\usepackage{booktabs}
\usepackage{multirow}
\usepackage{adjustbox}
\usepackage{cite}
\usepackage{amsmath}
\usepackage[normalem]{ulem}
\usepackage{soul} 
\usepackage{graphicx}
\begin{document}
\title{Online Regime-aware Calibration for Black-box Social Simulators \\via Posterior-assisted Evolutionary Dynamic Optimization}

\author{Peng~Yang,~\IEEEmembership{Senior Member,~IEEE},
        Zhenhua~Yang,
        Boquan~Jiang,
        Chenkai~Wang,
        Ke~Tang,~\IEEEmembership{Fellow,~IEEE},
        and~Xin~Yao,~\IEEEmembership{Fellow,~IEEE}%
\thanks{Peng Yang, Zhenhua Yang, Boquan Jiang, Chenkai Wang, and Ke Tang (\textit{corresponding author}) are with the Guangdong Provincial Key Laboratory of Brain-Inspired Intelligent Computation, Southern University of Science and Technology, Shenzhen 518055, China. Peng Yang, Zhenhua Yang, Boquan Jiang, and Ke Tang are also with the Department of Computer Science and Engineering, and Peng Yang and Chenkai Wang are also with the Department of Statistics and Data Science. Zhenhua Yang is also with Zhongguancun Academy, Beijing 100094, China. (e-mail: \{yangp, tangk3\}@sustech.edu.cn; \{12332460, wangck2022\}@mail.sustech.edu.cn; s-yzh24@bza.edu.cn).}%
\thanks{Xin Yao is with the School of Data Science, Lingnan University, Hong Kong SAR, China (e-mail: xinyao@ln.edu.hk).}%
}

\maketitle

\begin{abstract}

Evolutionary dynamic optimization (EDO) commonly assumes that environmental changes can be detected from fitness variations and handled through random re-initialization, historical solutions, or learned transition patterns. Online calibration of black-box simulators introduces a different setting, where the dynamic objective is induced by sequential observations and a changing calibration window, rather than being controlled by explicit variables. Fitness variations therefore cannot be directly attributed to regime changes, while the unknown relationship between successive regimes limits conventional adaptation.
We formulate this setting as an observation-driven dynamic optimization problem and propose PosEDO, which augments fitness-based EDO with an observation-conditioned parameter-space signal. 
PosEDO learns this signal online as a posterior distribution over simulator parameters from parameter–trajectory pairs generated during evolutionary evaluation, using posterior shifts for change detection and posterior samples for population adaptation. 
The new evaluation records are further utilized for online posterior updating without additional simulator calls.
Experiments on nonstationary economic and financial simulators show that PosEDO improves calibration accuracy, optimization performance, and change-detection quality over representative EDO baselines.

\end{abstract}

\begin{IEEEkeywords}
Evolutionary Dynamic Optimization, Online Calibration, Observation-Driven Dynamic Optimization.
\end{IEEEkeywords}

\section{Introduction}
\label{sec:introduction}

Computational simulators have been widely used to study complex social systems, such as economic systems~\cite{ref1}, financial markets~\cite{ref-zjj,yang2024reducing}, transportation systems~\cite{ref41}, and epidemic spreading~\cite{ref5}. These simulators describe macroscopic system behaviors with time series data through parameterized microscopic mechanisms, such as individual decision rules, interaction structures, population composition, and external conditions~\cite{ref1}.
Let
\(\mathrm{M}(\bm{\theta},1,T)=\{\bm{s}_t\}_{t=1}^{T}\)
denote the \(T\)-step trajectory generated by a simulator
\(\mathrm{M}\) under parameter vector
\(\bm{\theta}\in\mathbb{R}^{d}\).
Given an observed trajectory
\(\{\hat{\bm{s}}_t\}_{t=1}^{T}\) from the target system,
simulator calibration seeks parameters that minimize a prescribed discrepancy
\(\mathrm{D}(\cdot,\cdot)\)
between the simulated and observed trajectories~\cite{ref8,ref9}. In this sense, the calibrated parameters provide a mechanism-level explanation of the observed historical behaviors of the target system through the simulator. A conventional static calibration problem can be  formulated as $\bm{\theta}^{*}=\arg\min_{\bm{\theta}}\mathrm{D}\left(\{\hat{\bm{s}}_t\}_{t=1}^{T},\mathrm{M}(\bm{\theta},1,T)\right)$.

Static calibration assumes that the entire observation sequence can be explained by one fixed parameter setting. This assumption can be restrictive for real social systems, in which the underlying mechanisms may change over time~\cite{ref13}. Consequently, the observations collected before and after the change may no longer admit the same simulator-parameter explanation. Calibrating a single parameter over the entire sequence may thus mix incompatible observations and produce a parameter that explains neither period well. 

We therefore view the target system as evolving through a sequence of latent regimes. In this work, a regime is not defined merely as a statistically homogeneous segment of observations. Instead, it denotes an interval over which the observations admit a relatively stable explanation through a common simulator parameter.
Suppose that the current observation belongs to the $i$-th latent regime, whose unknown starting time is $\tilde{T}_i$. At the current time $T_c$, the ideal regime-level calibration objective is $\bm{\theta}_i^{*}(T_c)=\arg\min_{\bm{\theta}}\mathrm{D}\left(\{\hat{\bm{s}}_t\}_{t=\tilde{T}_i}^{T_c},\mathrm{M}(\bm{\theta},\tilde{T}_i,T_c)\right)$. The difficulty is that mechanism changes are not directly observable, and hence \(\tilde{T}_i\) is unknown. As new observations arrive, the calibrator must determine whether they remain compatible with the parameter explanation established for the current regime. If not, it must identify a new regime boundary, update the calibration window, and redirect the parameter search toward the new regime.

Several lines of work relate to this problem. Online parameter estimation and system identification~\cite{ref15,he2021} recursively update parameter estimates from each sequential observation. However, they typically track a time-varying parameter without explicitly identifying regime boundaries or determining which observations should share a common simulator parameter. Such boundaries are essential in our setting because they determine which observations enter the calibration objective. Bayesian online change-point detection identifies regime boundaries by evaluating whether new observations remain consistent with an assumed probabilistic model~\cite{alami2023bayesian,altamirano2023robust}. However, such methods rely on an explicit likelihood and are not directly applicable to black-box simulators.
Online learning with concept drift~\cite{lu2018learning} is also closely related, particularly in unsupervised settings where changes are detected from observations or their representations without regime labels~\cite{8246564,hinder2024oneA}. Its goal, however, is to preserve predictive performance under data-stream nonstationarity, without requiring the learned model parameters to represent the underlying mechanism. Consequently, a change in the data stream may justify updating a predictive model, but does not necessarily imply that the simulator parameters explaining the system have changed. Online regime-aware calibration instead asks whether newly arrived observations remain compatible with the simulator-parameter explanation of the current regime.

The above formulation is more directly related to Evolutionary Dynamic Optimization (EDO), which studies optimization problems whose objectives or constraints change over time~\cite{ref16,ref18}. 
We formulate online regime-aware calibration as an observation-driven DOP because its calibration objective is induced by sequentially received observations and the detected calibration window, rather than by an explicitly specified environment variable~\cite{ref16}. 
The key difficulty lies in the fact that the evolution of the calibration/fitness landscape is only indirectly observed through incoming observations. Even within a fixed regime, newly arrived observations introduce stochastic variations into the fitness landscape, whose magnitude and distribution are generally unknown. When a true regime change occurs, the relationship between the previous and new landscapes is also unknown because the data-generating process is nonstationary. Therefore, the observation-driven objective makes both change detection and population adaptation of conventional EDO unreliable or costly, as discussed below.

Conventional EDO methods commonly detect changes by monitoring the fitness variation~\cite{ref20,ref21}.
The fitness of a solution can vary simply because new observations come into the objective, even when the underlying mechanism remains unchanged. 
Population-level fitness analysis can reduce this sensitivity, but obtaining a comparable signal across observation windows generally requires re-evaluating the same set of parameters, which will incur substantial additional expensive simulator calls. 
After a change is detected, advanced methods typically adapt the population through reuse of historical solutions~\cite{ref22}, or prediction and transfer based on previous environments~\cite{ref45,ref48}. 
Unfortunately, these methods mostly rely on the assumption that the previous and new landscapes share similar patterns, which may be unreliable in the observation-driven tasks. Consequently, adaptation may still have to rely on generic random re-initialization~\cite{ref44}, which may cost more simulator calls to recover the search in the new regime.

Since the same incoming observations influence both the calibration objective and the suitability of simulator parameters, online calibration provides an opportunity to use a shared observation-conditioned parameter representation to assist both tasks.
Inspired by this, we introduce an observation-conditioned posterior distribution over simulator parameters. The ideal posterior
\(p(\bm{\theta}|\hat{\bm{s}})\)
assigns higher probability to parameters whose simulated outputs are more compatible with the current observation window under the calibration discrepancy. It thus represents the relative compatibility between the current observations and alternative simulator parameters. A substantial posterior shift provides evidence that this observation--parameter relation may have changed, while samples from the updated posterior can serve as promising candidates for population adaptation. The same posterior signal can therefore be reused across both detection and adaptation.

Because real observations are not readily accompanied by simulator parameters or regime labels, this posterior cannot be directly supervised from the target system. We instead sample candidate parameters
\(\bm{\theta}\), obtain the simulated data via simulation $\mathrm{M}(\bm{\theta})=s$,
and use the resulting parameter--trajectory pairs to train a conditional estimator
\(q_{\phi}(\bm{\theta}|\bm{s})\) parameterized by $\phi$.
The learned estimator is then applied to real observations through
\(q_{\phi}(\bm{\theta}|\hat{\bm{s}})\).

Posterior learning naturally integrates with population-based evolutionary search. The parameter--trajectory records generated during evolutionary evaluation are reused to train the posterior in an online manner without additional simulator calls, while posterior inference provides change evidence and candidate parameters for population adaptation. This bidirectional interaction forms the proposed posterior distribution-assisted EDO framework, termed PosEDO, implemented using a flow-based neural density estimator~\cite{papamakarios2017masked}.

The main contributions of this paper are threefold.
\begin{itemize}
    \item We formulate online regime-aware calibration of black-box social simulators as an observation-driven DOP, giving rise to new challenges for conventional EDOs.
    
    \item We propose PosEDO, a posterior-assisted EDO framework that learns an observation-conditioned parameter distribution online from searched parameter--trajectory pairs and for change detection and population adaptation.
    
    \item We conduct systematic experiments on nonstationary economic and financial simulators, demonstrating improved calibration, online optimization, and change-detection performance, with additional analyses of posterior-guided adaptation and higher-dimensional settings.
\end{itemize}

The remainder of this paper is organized as follows. Section II reviews related work. Section III formulates the problem. Section IV introduces the proposed PosEDO framework. Sections V--VII report the experimental results. Section VIII concludes this paper.

\section{Related Work}

\subsection{Simulator Calibration and Parameter Estimation}

Simulator calibration estimates the parameters of a given simulator by minimizing the discrepancy between simulated and observed data~\cite{ref8,ref9}. Existing methods can be broadly divided into point estimation and distribution estimation. Point-estimation methods search for a parameter setting that minimizes a prescribed discrepancy~\cite{ref8,ref36}, whereas distribution-estimation methods infer a distribution of parameters consistent with the observations~\cite{ref33,ref30}. For black-box simulators whose gradients or likelihoods are unavailable, calibration is commonly addressed using derivative-free optimization, including evolutionary algorithms~\cite{ref9,refyx2001}. Surrogate-assisted methods further reduce expensive simulator evaluations by approximating either the simulator or the parameter-fitness mapping~\cite{ref38,refyx2024}.

System identification and online parameter estimation are closely related. System identification infers system dynamics and parameters from sequential observations, typically under an explicitly specified model structure~\cite{billings2013nonlinear,yu2025review}. Such structural assumptions are often unavailable for complex black-box social simulators. Online parameter estimation focuses on updating current estimates as new observations arrive~\cite{ref15,he2021}, but usually maintains a continuously evolving estimate rather than explicitly identifying distinct mechanism regimes.

Simulation-based Bayesian Inference provides a probabilistic alternative for likelihood-free parameter inference~\cite{ref33,ref30}. Neural posterior estimation learns a conditional density \(q_\phi(\bm{\theta}|\bm{s})\) from simulator-generated parameter-trajectory pairs and enables efficient inference for new observations~\cite{ref30,ref35}. These methods provide the technical basis for the posterior estimator used in this work. However, existing methods mainly consider static or fixed-target inference and do not study how an amortized posterior can be updated and reused within an online evolutionary calibration process. Bayesian online change-point detection is an exception, as it jointly infers regime boundaries and segment-specific parameters by recursively updating their posterior distributions as observations arrive~\cite{alami2023bayesian,altamirano2023robust}. However, they require a predefined probabilistic model with a tractable likelihood, which is generally unavailable for black-box simulators.

\subsection{Online Learning with Concept Drift}

Online learning with concept drift studies predictive modeling under nonstationary data streams~\cite{lu2018learning,hinder2024oneA}. When labels are available, drift can be detected by monitoring prediction errors, losses, or other performance measures~\cite{hinder2024oneA,hinder2024oneB}. When labels are unavailable, unsupervised methods compare the distributions of observations or learned representations across time~\cite{hinder2024oneA,greco2025unsupervised}, using signals such as two-sample statistics or reconstruction errors~\cite{hinder2024oneA}.

These methods provide useful tools for detecting nonstationarity in sequential data. However, their detection and adaptation objectives are typically defined for maintaining predictive performance or characterizing observation-level drift~\cite{lu2018learning,hinder2024oneA}. They do not directly determine whether a sequence of observations should continue to share the same simulator-parameter explanation or how the simulator parameter search should be redirected after a change.

\subsection{Evolutionary Dynamic Optimization}

Evolutionary Dynamic Optimization (EDO) studies optimization problems whose objective functions, constraints, or fitness landscapes change over time~\cite{ref16,ref18,6148271}, with EAs. Existing methods mainly address two issues, i.e., change detection and population adaptation. 

Change detection aims to identify whether the current optimization environment has changed. A widely used strategy is to re-evaluate selected solutions, and detect changes by checking whether their fitness values have changed compared with their previously recorded values~\cite{ref21,ref42}. These selected solutions can be predefined detectors or sentinel solutions, elite individuals, the best-so-far solution, or representative individuals from the population~\cite{ref16}. Population-level methods further monitor statistics such as best fitness, average fitness, fitness variance, or fitness distribution to detect abrupt changes in the search landscape~\cite{ref21,nasiri2016history,niu2019bacterial}. These methods are effective when environmental changes are directly reflected by changes in fitness values or statistics.

After a change is detected, the algorithm needs to adapt the population to the new environment. Early adaptation strategies mainly focus on restoring diversity, for example by partially or fully re-initializing the population, increasing mutation strength, or adjusting control parameters to enlarge the exploration range~\cite{ref16,ref44}. Another line of work exploits historical search experience. Archive-based or memory-based methods store previously good solutions and reuse them when similar environments reappear~\cite{ref20,ref22}. Prediction-based methods estimate the direction or pattern of environmental changes and generate candidate solutions for future environments~\cite{ref24,ref25}. More recently, transfer optimization has been introduced to reuse knowledge from previous environments for population adaptation in the current environment~\cite{ref25-3}. Knowledge-transfer methods have also been developed for dynamic multi-objective optimization with a changing number of objectives, where populations or Pareto sets are adapted across changing environments~\cite{ruan2024knowledge,ruan2024learning}. In addition, generative model-based methods, such as diffusion-based prediction, have been explored to model transition patterns and generate promising candidates for new environments~\cite{ref25-1}.

These methods are generally designed for environments whose changes are reflected in the parameter-fitness mapping and whose historical landscapes may contain reusable transition information~\cite{ref18,ref23}. Online regime-aware calibration differs because this mapping is induced by a sequentially updated observation window. Consequently, the reliability and cost of fitness-based detection, as well as the ability of conventional adaptation methods to exploit current observations, require further examination. Section III formalizes these differences and analyzes their implications for online calibration.

\section{Problem Definition and Challenges}
\label{sec:problem}
This section formulates online regime-aware calibration as an observation-driven DOP and analyzes the challenges it brings to change detection and population adaptation. 

\subsection{Observation-Driven DOP}

As discussed in Section I, the ground-truth starting time $\tilde{T}_i$ of the current regime is unobservable, hence the previously introduced ideal regime-level calibration objective cannot be directly evaluated. 
We therefore define a practical objective by replacing the unknown regime start $\tilde{T}_i$ with the latest detected change point $T_i$ by the online calibrator.

Let $\{\hat{\bm{s}}_t\}$ denote the incoming observation stream, where each $\hat{\bm{s}}_t\in\mathbb{R}^{L\times m}$ is the minimum observation unit. Here, $L$ is the length of the observation window and $m$ is the dimension of observed variables.
Let the current calibration window be $[T_i,T_c]$.
Given a simulator parameter $\bm{\theta}$, the practical fitness used for online regime-level calibration is defined as
\begin{equation}
f(\bm{\theta};\{\hat{\bm{s}}_t\},T_i,T_c)
=
\mathrm{D}
\left(
\{\hat{\bm{s}}_t\}_{t=T_i}^{T_c},
\mathrm{M}(\bm{\theta},T_i,T_c)
\right),
\label{eq:practical_fitness}
\end{equation}
where $\mathrm{M}(\bm{\theta},T_i,T_c)=\{\bm{s}_t\}_{t=T_i}^{T_c}$ indicates the simulated trajectory $\{\bm{s}_t\}_{t=T_i}^{T_c}$ generated by simulator $M$ under $\bm{\theta}$ over the same interval, and $\mathrm{D}(\cdot,\cdot)$ measures the discrepancy between $\{\bm{s}_t\}_{t=T_i}^{T_c}$ and $\{\hat{\bm{s}}_t\}_{t=T_i}^{T_c}$. The corresponding practical online calibration objective at time $T_c$ is then
\begin{equation}
\bm{\theta}^{*}(T_i,T_c)
=
\arg\min_{\bm{\theta}}
f(\bm{\theta};\{\hat{\bm{s}}_t\},T_i,T_c).
\label{eq:practical_objective}
\end{equation}

This formulation differs from the conventional DOP setting, which is often written as
\begin{equation}
\bm{\theta}^{*}(\alpha_i)
=
\arg\min_{\bm{\theta}}
f(\bm{\theta};\alpha_i),
\label{eq:traditional_dop}
\end{equation}
where $\alpha_i$ denotes an environment variable that determines the objective function or constraints of the $i$-th environment. In such problems, the optimizer tracks changing optima as $\alpha_i$ changes over time. 

In online regime-aware calibration, the dynamic objective is not indexed by an explicit environment variable $\alpha_i$. Instead, it is induced by the observation stream $\{\hat{\bm{s}}_t\}$ and the current calibration window $[T_i,T_c]$, which determine how each simulator parameter is evaluated. We therefore call it an observation-driven dynamic optimization problem.

\subsection{Challenges to Conventional EDO}
In conventional EDO, the environment variable $\alpha_i$ determines the dynamics of the mapping from candidate parameters to fitness values, i.e., $\bm{\theta}\mapsto f(\bm{\theta};\alpha_i)$. Therefore, change detection and adaptation can be built on assumptions about how this mapping changes across environments, such as abrupt shifts, smooth movements~\cite{ref18,ref24}, recurring environments~\cite{ref22}, or transferable patterns between historical and current landscapes~\cite{ref23,ref25-3}. In online regime-aware calibration, however, the mapping from $\bm{\theta}$ to fitness is induced by the current observation window through $f(\bm{\theta};\{\hat{\bm{s}}_t\},T_i,T_c)$. Since this window evolves as observations arrive and as $T_i$ is updated, changes in the parameter-fitness relation may be caused by ordinary window updates or by true mechanism changes. This is the shared source of the following two challenges.

First, fitness-based change detection is ambiguous and potentially expensive. For a single fixed parameter \(\bm{\theta}\), the fitness variation
$\Delta f_c(\bm{\theta})=\left|f(\bm{\theta};\{\hat{\bm{s}}_t\},T_i,T_c)
-f(\bm{\theta};\{\hat{\bm{s}}_t\},T_i,T_c-1)\right|$
may be nonzero simply because the newly received observation updates the calibration window, even when the underlying regime remains unchanged. Hence, single-solution fitness variation can produce false alarms.

Population-level fitness analysis can reduce the sensitivity of a single detector by comparing multiple parameters or aggregate fitness statistics. However, a meaningful comparison across observation windows requires re-evaluating the same parameter set under both windows. This introduces additional simulator calls at every observation time. Instead, if the fitness distributions of the populations generated in successive iterations are directly compared, which can avoid re-evaluation, the observed difference reflects both the change in observations and the movement of the population, making the source of the variation difficult to identify.

Second, population adaptation receives limited guidance from the detected change. Conventional methods typically restore diversity through re-initialization or relocation~\cite{ref44}, reuse historical solutions through memory or archive mechanisms~\cite{ref20,ref22}, or predict and transfer candidate solutions from previous environmental transitions~\cite{ref23,ref24}.
These strategies can restart or redirect the search, but the change detector generally outputs only a change indication and provides little information about which parameters are compatible with the current observations. Historical reuse and transition prediction are useful when environments recur or optima follow regular trajectories, but may be less effective for unseen regimes without such structure. Random re-initialization avoids these assumptions but can consume many expensive simulator evaluations to recover the search. In contrast, online calibration provides an opportunity to use information from the current observations to generate more informative candidate parameters after a detected change, without the re-evaluation of parameters.

These challenges motivate a shared observation-conditioned representation over simulator parameters. Such a representation can provide a population-independent signal for comparing consecutive observation windows and can also generate candidate parameters for adaptation after a detected change, both reliably and at low computational cost. The posterior distribution introduced in the next section is designed to serve these two purposes.

\section{The Proposed Method}
\label{sec:method}

To address the above challenges, we propose a posterior distribution-assisted EDO framework, termed PosEDO. Its key idea is to learn an observation-conditioned distribution over simulator parameters and reuse it across change detection and population adaptation. Let \(q_\phi(\bm{\theta}|\hat{\bm{s}})\) denote the learned conditional estimator. At each online step, parameter-trajectory records generated by evolutionary search are used to incrementally train \(q_\phi\). The updated estimator is then used to assess changes in the observation-parameter relation and, after a detected change, to generate candidate parameters for the new population.

\subsection{Posterior-assisted Change Detection and Adaptation}

For change detection, the posterior helps distinguish real distributional shifts from false alarms caused by an updated observation window. If no mechanism change occurs, newly arrived data should still be generated by the same underlying regime, and the corresponding posterior distribution should remain relatively stable. In contrast, a substantial change in the posterior indicates that the newly observed data are better explained by a different parameter, suggesting a possible mechanism change of the target system.

Algorithm~\ref{alg:change-detection} describes the change detection procedure.
At time \(T_c\), we compare the posterior conditioned on the historical observations in the current regime, \(q_\phi(\bm{\theta}| \{\hat{\bm{s}}_t\}_{t=T_i}^{T_c-1})\), with the posterior conditioned on the newly arrived observation, \(q_\phi(\bm{\theta}| \hat{\bm{s}}_{T_c})\). Their discrepancy is measured by the Monte Carlo Kullback–Leibler (KL) divergence:
\begin{equation}
\label{eq:kl}
\resizebox{0.91\linewidth}{!}{$\displaystyle
\mathrm{D}_{\mathrm{KL}}
\approx
\frac{1}{N}\sum_{n=1}^{N}
\log
\frac{
q_\phi(\bm{\theta}_n \mid \hat{\bm{s}}_{T_c})
}{
q_\phi\!\left(
\bm{\theta}_n
\mid
\{\hat{\bm{s}}_t\}_{t=T_i}^{T_c-1}
\right)
},
\quad
\bm{\theta}_n
\sim
q_\phi(\bm{\theta}\mid\hat{\bm{s}}_{T_c})
$}
\end{equation}

\noindent where \(\{\bm{\theta}_n\}_{n=1}^{N}\) are sampled parameters from $q_\phi(\bm{\theta}|\hat{\bm{s}}_{T_c})$ for Monte Carlo approximation. If \(\mathrm{D}_{\mathrm{KL}}\geq \varepsilon\), the new observation is regarded as indicating a real system change, and \(T_c\) is recorded as a new change point. 

\begin{algorithm}[tbp]
\caption{ChangeDetection(\(q_\phi, \{\hat{\bm{s}}_t\}_{t=T_i}^{T_c}, \varepsilon\))}
\label{alg:change-detection}
\begin{algorithmic}[0]
\STATE \hspace{-1em} \textbf{Input:} posterior \(q_\phi\); observed data since the latest detected change point \(\{\hat{\bm{s}}_t\}_{t=T_i}^{T_c}\); threshold \(\varepsilon\).
\STATE \hspace{-1em} \textbf{Output:} change flag \(flag\).
\end{algorithmic}
\begin{algorithmic}[1]
\STATE Initialize \(flag \gets \FALSE\);
\STATE Compute \(\mathrm{D}_{\mathrm{KL}}\) between \(q_\phi(\bm{\theta}| \{\hat{\bm{s}}_t\}_{t=T_i}^{T_c-1})\) and \(q_\phi(\bm{\theta}| \hat{\bm{s}}_{T_c})\) using Eq.~\eqref{eq:kl};
\IF{\(\mathrm{D}_{\mathrm{KL}} \geq \varepsilon\)}
    \STATE \(flag \gets \TRUE\);
\ENDIF
\end{algorithmic}
\end{algorithm}

For environmental adaptation, the posterior provides a computationally cheap way to generate promising candidate parameters for the new environment. As summarized in Algorithm~\ref{alg:environment-adaptation}, once a change is detected at \(T_c\), the old population is discarded and a new population \( \bm{\Theta} = \{\bm{\theta}_j\}_{j=1}^{\lambda}\) is sampled from \(q_\phi(\bm{\theta}| \hat{\bm{s}}_{T_c})\). 
To avoid excessive concentration, we first draw a candidate set \(\{\bm{\theta}_n\}_{n=1}^{N}\) from the posterior and then greedily select \(\lambda\) diverse individuals by maximizing their pairwise Euclidean distances, where \(\lambda\) is the population size.

\begin{algorithm}[tbp]
\caption{Adaptation(\(q_\phi\), \(\hat{\bm{s}}_{T_c}\), \(\lambda\), \(N\))}
\label{alg:environment-adaptation}
\begin{algorithmic}[0]
\STATE \hspace{-1em} \textbf{Input:} posterior \(q_\phi\); current observation \(\hat{\bm{s}}_{T_c}\); population size \(\lambda\); candidate sample size \(N\).
\STATE \hspace{-1em} \textbf{Output:} adapted population \(\bm{\Theta}=\{\bm{\theta}_j\}_{j=1}^{\lambda}\).
\end{algorithmic}
\begin{algorithmic}[1]
\STATE Draw candidate samples \(\{\bm{\theta}_n\}_{n=1}^{N}\sim q_\phi(\bm{\theta}| \hat{\bm{s}}_{T_c})\);
\STATE Initialize \(\bm{\Theta}\gets \{\bm{\theta}_1\}\), where \(\bm{\theta}_1\sim \mathrm{Uniform}(\{\bm{\theta}_n\}_{n=1}^{N})\);
\FOR{\(j=2\) to \(\lambda\)}
    \STATE
    \(\bm{\theta}_j =
    \operatorname*{argmax}_{\bm{\theta}\in \{\bm{\theta}_n\}_{n=1}^{N}\setminus \bm{\Theta}}
    \sum_{\bm{\theta}'\in \bm{\Theta}}
    \|\bm{\theta}-\bm{\theta}'\|_2\);
    \STATE \(\bm{\Theta}\gets \bm{\Theta}\cup \{\bm{\theta}_j\}\);
\ENDFOR
\end{algorithmic}
\end{algorithm}

\subsection{Flow-based Posterior Distribution Estimator}

We estimate the posterior distribution using a conditional masked autoregressive flow (MAF) \cite{ref30}, parameterized as \(q_\phi(\bm{\theta}| \{\bm{s}_t\})\). The MAF defines an invertible transformation \(f_\phi\) from a simple latent distribution to the parameter space:
$\bm{\theta}=f_\phi(\bm{z};\{\bm{s}_t\}), \quad \bm{z}\sim \mathcal{N}(\bm{0},\mathbf{I}),$
where the simulated or observed data \(\{\bm{s}_t\}\) are used as the condition.

In our implementation, \(f_\phi\) consists of five masked autoregressive affine transformation layers interleaved with five random permutation layers. For each transformation layer \(l\), the \(r\)-th dimension is computed as
\[
\bm{\theta}_r^{(l)}
=
\mu_r^{(l)}(\bm{\theta}_{1:r-1}^{(l-1)},\mathrm{R}(\{\bm{s}_t\}))
+
\sigma_r^{(l)}(\bm{\theta}_{1:r-1}^{(l-1)},\mathrm{R}(\{\bm{s}_t\}))
\cdot
\bm{\theta}_r^{(l-1)},
\]
where \(\mu_r^{(l)}(\cdot)\) and \(\sigma_r^{(l)}(\cdot)\) are implemented by masked multilayer perceptrons (MLPs), and $\bm{\theta}^{(0)}=\bm{z}$. Each MLP consists of two fully connected layers with 50 hidden units and ReLU activations.
The conditioning data \(\{\bm{s}_t\}\) are first represented by a summary function \(\mathrm{R}(\cdot)\) into an embedded vector of dimension \(k\), i.e., \(\mathrm{R}(\{\bm{s}_t\}) \in \mathbb{R}^k\), 
which serves as an additional input to each autoregressive transformation to incorporate contextual information from the observed data. 
In practice, we define $\mathrm{R}(\cdot)$ using the commonly adopted $k=9$ length-invariant summary statistics of time series data, i.e., the mean, median, variance, range, minimum, maximum, skewness, kurtosis, and interquartile range of a given $\{\bm{s}_t\}$.

Given a training pair \(\langle \bm{\theta},\{\bm{s}_t\}\rangle\), the inverse transformation \(f_\phi^{-1}\) maps \(\bm{\theta}\) back to \(\bm{z}\), allowing the conditional log-likelihood to be computed as
\begin{equation}
\label{eq:log-likelihood}
\resizebox{\linewidth}{!}{$\displaystyle
\log q_\phi\!\left(
\bm{\theta}
\mid
\{\bm{s}_t\}
\right)
=
\log p(\bm{z})
+
\log
\left|
\det
\left(
\frac{
\partial f_\phi^{-1}
\!\left(
\bm{\theta};
\{\bm{s}_t\}
\right)
}{
\partial \bm{\theta}
}
\right)
\right|
$}
\end{equation}

At the beginning of online calibration, \(q_\phi\) is randomly initialized. Subsequently, \(q_\phi\) is trained online by minimizing the negative conditional log-likelihood of a given dataset $\mathcal{D}$:
\begin{equation}
\label{eq:train_loss}
\mathcal{L}(\phi)
=
-\frac{1}{B}
\sum_{m=1}^{B}
\log q_\phi(\bm{\theta}_m| \{\bm{s}_t\}_m),
\end{equation}
where \(B\) is the mini-batch size. Training is terminated early if the loss shows no improvement for $E_{max}$ consecutive epochs, 
to prevent overfitting and reduce computational costs.
The dataset $\mathcal{D}$ is collected from the parameter--trajectory pairs generated during evolutionary search, hence no additional simulator calls is needed.
The detailed algorithm for training the flow-based posterior is presented in Algorithm~\ref{alg:training}.

\begin{algorithm}[tbp]
\caption{Training(\(q_\phi\), \(\mathcal{D}\), \(E_{\max}\))\cite{ref30}}
\label{alg:training}
\begin{algorithmic}[0]
\STATE \hspace{-1em} \textbf{Input:} posterior \(q_\phi\); training dataset \(\mathcal{D}=\{\langle \bm{\theta}_m,\{\bm{s}_t\}_m\rangle\}\); early-stopping epoch \(E_{\max}\).
\STATE \hspace{-1em} \textbf{Output:} updated posterior \(q_\phi\).
\end{algorithmic}
\begin{algorithmic}[1]
\STATE Initialize \(e\gets 0\), \(\mathcal{L}_{best}\gets \infty\);
\WHILE{\(e<E_{\max}\)}
    \FOR{each mini-batch from \(\mathcal{D}\)}
        \STATE Compute \(\mathcal{L}(\phi)\) using Eq.~\eqref{eq:train_loss};
        \STATE Update \(\phi \leftarrow \phi-\eta\nabla_\phi \mathcal{L}(\phi)\);
    \ENDFOR
    \IF{\(\mathcal{L}(\phi)<\mathcal{L}_{best}\)}
        \STATE \(\mathcal{L}_{best}\gets \mathcal{L}(\phi)\), \(e\gets 0\);
    \ELSE
        \STATE \(e\gets e+1\);
    \ENDIF
\ENDWHILE
\end{algorithmic}
\end{algorithm}

\subsection{Optimization and Training Data Generation}

\begin{algorithm}[tbp]
\caption{Optimization(\(\bm{\Theta}\), \(I_{\max}\), \(\lambda\), \(T_i\), \(T_c\))}
\label{alg:optimization}
\begin{algorithmic}[0]
\STATE \hspace{-1em} \textbf{Input:} population \(\bm{\Theta}\); maximum iterations \(I_{\max}\); population size \(\lambda\); latest detected change point \(T_i\); current time \(T_c\).
\STATE \hspace{-1em} \textbf{Output:} best-found parameter \(\bm{\theta}^{*}\); updated population \(\bm{\Theta}\); online training dataset \(\mathcal{D}\).
\end{algorithmic}
\begin{algorithmic}[1]
\STATE Initialize \(iter\gets 1\), \(\mathcal{D}\gets \emptyset\);
\WHILE{\(iter\leq I_{\max}\)}
    \FOR{each individual \(\bm{\theta}_j\in \bm{\Theta}\)}
        \STATE \(\{\bm{s}_t\}_{t=T_i}^{T_c}\gets
        \mathrm{M}(\bm{\theta}_j,T_i,T_c)\);
        \STATE Evaluate fitness using Eq.~\eqref{eq:fitness};
        \STATE
        \(\mathcal{D}
        \gets
        \mathcal{D}
        \cup
        \left\{
        \left\langle
        \bm{\theta}_j,
        \{\bm{s}_t\}_{t=T_i}^{T_c}
        \right\rangle
        \right\}\);
    \ENDFOR
    \STATE Keep the best-found parameter as \(\bm{\theta}^{*}\);
    \STATE Update \(\bm{\Theta}\) using evolutionary operators;
    \STATE \(iter\gets iter+1\);
\ENDWHILE
\end{algorithmic}
\end{algorithm}

At time \(T_c\), the population \(\bm{\Theta}\) is evolved for \(I_{\max}\) iterations to search for \(\bm{\theta}_i^{*}\), short for \(\bm{\theta}^{*}(T_i,T_c)\), by solving Eq.~\eqref{eq:practical_objective}. Any evolutionary operator can be used in this optimization module, including the multi-population management \cite{ref27, ref47} commonly adopted in EDOs.
To implement the discrepancy metric, this work adopts the method of simulated moments \cite{ref9,ref36}:
\begin{equation}
\label{eq:fitness}
\resizebox{\linewidth}{!}{$\displaystyle
\mathrm{D}
\left(
\{\hat{\bm{s}}_t\}_{t=T_i}^{T_c},
\{\bm{s}_t\}_{t=T_i}^{T_c}
\right)
=
\frac{1}{k}
\left\|
\mathrm{R}
\left(
\{\hat{\bm{s}}_t\}_{t=T_i}^{T_c}
\right)
-
\mathrm{R}
\left(
\{\bm{s}_t\}_{t=T_i}^{T_c}
\right)
\right\|_2^2
$}
\end{equation}

During optimization, the simulator naturally generates parameter--trajectory pairs. We collect these pairs to form an online training dataset
\(\mathcal{D}=\left\{\left\langle\bm{\theta}_j,\mathrm{M}(\bm{\theta}_j,T_i,T_c)\right\rangle\right\}\).
The size of the dataset is determined by the number of simulator calls performed during the optimization, normally \(\lambda\cdot I_{\max}\).
The posterior \(q_\phi\) is then trained online on \(\mathcal{D}\), enabling it to adapt to the local data distribution of the current online calibration process without additional simulator calls. The complete workflow is described in Algorithm~\ref{alg:optimization}.

\subsection{The PosEDO Framework}

The goal of PosEDO is to sequentially search one calibrated parameter
for each detected regime. Let \(i\) index the detected regimes and let
\(T_i\) denote the starting time of the current regime. At time \(T_c\),
PosEDO maintains the best-found parameter \(\bm{\theta}_i^*\) for the
current calibration window \([T_i,T_c]\). Once a new change is detected
at \(T_{i+1}\), this parameter is finalized and archived as the calibrated
parameter for the completed regime \([T_i,T_{i+1}-1]\). The resulting
set of regime-specific calibrated parameters is denoted by
\(\bm{S}=\{\bm{\theta}_i^*\}\).

As shown in Algorithm~\ref{alg:online-calibration}, the posterior
estimator \(q_\phi\) and the population \(\bm{\Theta}\) are randomly
initialized at the beginning of online calibration. For the first
observation, no historical observations are available for change
detection and thus $T_c \neq T_i$. PosEDO therefore directly performs optimization and uses the
generated parameter--trajectory pairs to train the posterior estimator.

For each subsequent observation \(\hat{\bm{s}}_{T_c}\), PosEDO first
determines whether it remains compatible with the current regime using
posterior-based change detection. If no change is detected, the current
regime start \(T_i\) and population \(\bm{\Theta}\) are retained, and
optimization is performed over the extended calibration window
\([T_i,T_c]\). If a change is detected, the best parameter of the
previous regime is archived, \(T_c\) is recorded as the starting time of
a new regime, and the regime index is updated. The posterior conditioned
on the new observation is then used to initialize a new population
through posterior-guided adaptation, after which optimization is
restarted using the new observation as the initial observation of the
new calibration window.
During optimization at each time step, the parameter--trajectory pairs
generated by evolutionary search are continuously provided to the online posterior-training process. The posterior estimator \(q_\phi\)
is therefore updated without requiring additional simulator calls. The updated posterior is
used for change detection and population adaptation when the next
observation arrives, while the best-found parameter is retained as the
current estimate \(\bm{\theta}_i^*\).

\begin{algorithm}[tbp]
\caption{The PosEDO Framework}
\label{alg:online-calibration}

\begin{algorithmic}[0]
\STATE \hspace{-1.32em} \textbf{Input:}
detection threshold \(\varepsilon\);
maximum iterations for one optimization \(I_{\max}\); training epoch for one online training \(E_{max}\);
population size \(\lambda\);
candidate sample size \(N\).
\STATE \hspace{-1.32em} \textbf{Output:}
calibrated parameters \(\bm{S}\);
detected change points \(\mathbf{T}\).
\end{algorithmic}

\begin{algorithmic}[1]
\STATE Randomly initialize posterior estimator \(q_\phi\);
\STATE Initialize
\(\bm{S}\gets\emptyset\),
\(\mathbf{T}\gets\emptyset\),
\(T_c\gets1\),
\(i\gets1\), and \(T_1\gets1\);
\STATE Randomly initialize population
\(\bm{\Theta}=\{\bm{\theta}_j\}_{j=1}^{\lambda}\);

\FOR{each observation \(\hat{\bm{s}}_{T_c}\) at time \(T_c\)}

    \IF{\(T_c\neq T_i\)}
        \IF{\(
            \mathrm{ChangeDetection}
            (q_\phi,
            \{\hat{\bm{s}}_t\}_{t=T_i}^{T_c},
            \varepsilon)\)}
            \STATE
            \(T_{i+1}\gets T_c\),
            \(\mathbf{T}\gets
            \mathbf{T}\cup\{T_{i+1}\}\);
            \STATE
            \(\bm{S}\gets
            \bm{S}\cup\{\bm{\theta}_i^*\}\);
            \(i\gets i+1\);
            \STATE
            \(\bm{\Theta}\gets
            \mathrm{Adaptation}
            (q_\phi,\hat{\bm{s}}_{T_c},\lambda,N)\);
        \ENDIF
    \ENDIF

    \STATE
    \(\{\bm{\theta}_i^*,\bm{\Theta},\mathcal{D}\}
    \gets
    \mathrm{Optimization}
    (\bm{\Theta},I_{\max},\lambda,T_i,T_c)\);

    \STATE
    \(q_\phi\gets
    \mathrm{Training}
    (q_\phi,\mathcal{D},E_{max})\);

    \STATE \(T_c\gets T_c+1\);
\ENDFOR

\end{algorithmic}
\end{algorithm}

\subsection{Computational Complexity Analysis}

At each observation time \(T_c\), the population contains \(\lambda\)
individuals and is evolved for \(I_{\max}\) iterations. The computational
cost of evolutionary optimization is
\(\mathcal{T}_{\mathrm{opt}}(T_c)
=\mathcal{O}(\lambda I_{\max}C_{\mathrm{sim}})\), where
\(C_{\mathrm{sim}}\) denotes the average cost of one simulation
call.
During optimization, the generated parameter--trajectory pairs are
provided to the posterior-training process. Under the basic
evaluation setting, \(\lambda I_{\max}\) pairs are generated at each
observation time. The computational workload of online posterior
training is
\(\mathcal{T}_{\mathrm{train}}(T_c)
=\mathcal{O}(E_{max}\lambda I_{\max}C_{\mathrm{train}})\),
where \(E_{max}\) denotes the online training epochs and
\(C_{\mathrm{train}}\) is the average per-sample training cost.
Change detection estimates the KL
divergence using \(N\) posterior samples, resulting in
\(\mathcal{T}_{\mathrm{cd}}(T_c)
=\mathcal{O}(NC_{\mathrm{post}})\), where \(C_{\mathrm{post}}\)
denotes the average cost of posterior sampling and density evaluation.
When a regime change is detected, posterior-guided adaptation draws
\(N\) candidate parameters and greedily selects \(\lambda\) diverse
individuals, with computational cost
\(\mathcal{T}_{\mathrm{ada}}
=\mathcal{O}(NC_{\mathrm{post}}+N\lambda^2)\).

Suppose that \(T\) observations arrive sequentially and \(K_d\) regime
changes are detected. The total computational workload is
\(\mathcal{T}_{\mathrm{PosEDO}}(T)
=
T\cdot\mathcal{T}_{\mathrm{opt}}(T_c)
+T\cdot\mathcal{T}_{\mathrm{train}}(T_c)
+(T-1)\cdot\mathcal{T}_{\mathrm{cd}}(T_c)
+K_d\mathcal{T}_{\mathrm{ada}}
=\mathcal{O}\Big(T\lambda I_{\max} C_{\mathrm{sim}}
 + T E_{\mathrm{max}}\lambda I_{\max} C_{\mathrm{train}} + T N C_{\mathrm{post}}
 + K_d(N C_{\mathrm{post}}+N\lambda^2)
 \Big)\).
Since simulator evaluations are usually much more expensive than posterior inference and neural-network updates, the practical dominant term is $\mathcal{T}_{\mathrm{PosEDO}}(T)\approx \mathcal{O}(T\lambda I_{\max} C_{\mathrm{sim}})$. Such complexity clearly show that PosEDO does not require additional simulation calls for change detection, adaptation or online training.

\section{Experimental Setup}
\label{sec:setup}

This section describes the benchmark simulators, compared methods,
evaluation metrics, and experimental configurations.

\subsection{Simulators and Benchmark Instances}

We consider two complementary economic and financial simulators, i.e.,
the Brock--Hommes (BH) model~\cite{ref2} for asset-price dynamics and
the Preis--Golke--Paul--Schneid (PGPS) model~\cite{ref48} for market
microstructure. Both generate price-related time series through
interactions among heterogeneous multi-agents, but differ substantially in
parameter dimensionality and behavioral complexity. Both models have
been frequently studied in static calibration settings~\cite{ref9,ref-rjj}.

\subsubsection{The BH Model}
The BH model describes nonlinear asset-price dynamics generated by
heterogeneous expectations. We calibrate two parameters,
\(\bm{\theta}=[g_2,b_2]\), where \(g_2\in[0.5,1]\) controls the
trend-following intensity and \(b_2\in[0,1]\) controls the expectation
bias of trend-following agents. These parameters affect aggregate demand
and subsequent price dynamics, making BH a low-dimensional diagnostic
benchmark for online calibration.
We construct nine BH instances under three change-frequency settings
with \(2\), \(4\), and \(8\) ground-truth change points, corresponding
to \(3\), \(5\), and \(9\) latent regimes, respectively. Each setting
contains three independently generated instances. Every instance
consists of \(30\) sequential observations, each represented by a
univariate time series of size \(L\times m=50\times1\). Regime changes
are generated by switching the simulator parameters at predefined
change points that are hidden from the online calibrator.

\subsubsection{The PGPS Model}

The PGPS model is an agent-based limit-order-book (LOB) simulator containing
liquidity providers and liquidity takers who observe the LOB of current market and submit trading various orders with different strategies. In our experiments, the market
contains \(125\) agents of each type. The calibration target is the
mid-price \(p_{\mathrm{mid}}(t)=[p_a(t)+p_b(t)]/2\), where \(p_a(t)\)
and \(p_b(t)\) denote the best ask and bid prices, respectively. The
original model contains six calibrated parameters,
\(\bm{w}=[\alpha,\mu,\delta,\Delta S,\lambda_0,C_\lambda]\), which
control the probability of limit-order submission, the probability of
market-order submission, the probability of order cancellation, the
mean-reversion step, the initial order-book depth, and the step size for
varying the order-book depth, respectively. Compared with BH, PGPS
provides a more stochastic and interaction-intensive calibration
benchmark.

For the six-dimensional PGPS (PGPS-6D) benchmark, we construct nine instances.
Each instance consists of \(18\) sequential observation blocks, and
each block is a univariate mid-price time series of size
\(L\times m=200\times1\), yielding \(3600\) simulator time steps in
total. Each instance contains six latent regimes of three observation
blocks each. The five ground-truth changes therefore occur after blocks
\(3,6,9,12,\) and \(15\), corresponding to simulator time indices
\(600,1200,1800,2400,\) and \(3000\). These change points are hidden
from the online calibrator.

To evaluate scalability, we further extend PGPS to parameter dimensions
\(d\in\{12,24,36,48\}\) by introducing group-specific behavioral
parameters. The agents are evenly divided into \(d/3\) heterogeneous
groups, each governed by three parameters controlling the provider price
offset, provider submission frequency, and taker submission frequency.
The global parameters are fixed as
\(\lambda_0=100\), \(C_\lambda=10\), and \(\Delta S=0.001\).
For each dimensional setting, we construct three independently generated
instances. Each instance follows the same observation-stream structure
as the PGPS-6D benchmark. The same number of independent runs
and optimization settings are used across all dimensional settings.

\subsection{Compared Methods}

We compare PosEDO with EDO baselines constructed by combining
representative change-detection and population-adaptation mechanisms.
All methods use the same adaptive multi-population PSO (AMP/PSO)
backbone for evolutionary search ~\cite{ref27}, motivated by the multimodal nature of simulator
calibration~\cite{ref39}. The original AMP/PSO dynamically creates and
deletes subpopulations. To control the optimization budget, we disable
these operations and fix the total population size to \(\lambda=40\).
All methods also use the same optimization iterations and parameter
bounds. Therefore, their differences arise from change detection and
post-change population adaptation rather than from the underlying
optimizer.

We consider three change-detection mechanisms. First, detector-based detection (DBD) maintains a fixed set of detector solutions~\cite{ref20}. The number of detector solutions is set equal to the population size.
Before the normal optimization process at each new observation, these detector solutions are re-evaluated on both the previously unchanged observation window and the newly extended observation window. The average absolute fitness difference over all detectors is used as the change score. A change is declared when this score exceeds an adaptive threshold controlled by \(\alpha\), where the threshold is set according to the fitness-variation statistics observed in the previous unchanged window. These detector re-evaluations are included in the reported number of
function evaluations (NFE).
DBD is combined with random re-initialization, yielding DBD-Rand, and multiple values of \(\alpha\) are evaluated.

Following fitness-monitoring-based change detection in
EDO, we construct two simple baselines. FBCD is inspired by
methods that monitor elite fitness values to identify environmental
changes~\cite{ref21}, whereas MFBCD follows the use of
population-level average fitness as a change signal
~\cite{nasiri2016history,niu2019bacterial}.
Both methods follow an optimize-then-detect workflow. They first
optimize the parameters after a new observation arrives and then compare
the resulting best fitness or population mean fitness, respectively,
with the corresponding value from the most recent observation judged
to be stable. A change is declared when the monitored fitness fails to
improve. The current observation is then treated as the beginning of a
new regime, the population is adapted, and optimization is repeated for
the new calibration window. Consequently, both true and false detections
may introduce repeated optimization and additional simulator
evaluations.

Four population-adaptation strategies are considered. Random
re-initialization (Rand) discards the current population and uniformly
samples a new population within the parameter bounds. Archive-based
adaptation (Arch) stores elite solutions from previous observation
windows and uses selected archived elites and their perturbations to
initialize the new population~\cite{ref22}. Neural-network-based
information transfer (NNIT) follows~\cite{ref23} and learns a neural
mapping from historical optimization data to transform previously
obtained solutions into candidates for the current environment.
Data-driven evolutionary transfer initialization (DETOInit) uses
historical elite solutions and their estimated movement directions to
warm-start the population, following the transfer mechanism of
data-driven evolutionary optimization in dynamic environments~\cite{ref25-3}. Only its
initialization strategy is adopted, while the common AMP/PSO backbone is
retained. All history-based methods use only information available
before the current observation.

Combining FBCD and MFBCD with the four adaptation strategies yields
eight baselines: FBCD-Rand, FBCD-Arch, FBCD-NNIT, FBCD-DETOInit,
MFBCD-Rand, MFBCD-Arch, MFBCD-NNIT, and MFBCD-DETOInit. Together with
DBD-Rand, these methods form the main comparison set.

We additionally construct two PosEDO variants for controlled analyses.
PosEDO-CD retains posterior-based change detection but replaces
posterior-guided adaptation with random re-initialization, thereby enabling a controlled comparison with the other Rand-based change-detection methods. PosEDO-Ada is supplied with
the ground-truth change points and retains posterior-guided adaptation,
serving as an oracle-change reference. To compare adaptation mechanisms
independently of detection errors, Rand, Arch, NNIT, DETOInit, and
PosEDO-Ada are also evaluated under the same ground-truth change points.

\subsection{Performance Metrics}

We evaluate regime-level calibration accuracy, online convergence,
simulator-evaluation cost, and change-detection quality using
\(E_{\mathrm{MCE}}\), \(P_{\mathrm{CON}}\), NFE, and standard
change-point detection metrics, respectively.

\subsubsection{Mean Calibration Error}

The mean calibration error \(E_{\mathrm{MCE}}\) is evaluated through
post-hoc replay. For each independent algorithm run \(r\), the calibrated
parameters associated with the detected regimes are replayed \(K\) times,
and the resulting segment trajectories are combined into a complete
trajectory. The replay and observed trajectories are then partitioned
according to the \(C\) ground-truth regimes, which are used only for
evaluation.

Let \(\widehat{\mathcal{S}}_i\) and
\(\mathcal{S}_{r,k,i}\) denote the observed and replayed trajectories,
respectively, within the \(i\)-th ground-truth regime. Its scale-adjusted
replay error is
\begin{equation}
\label{eq:Ec}
e_{r,k,i}
=
\frac{
\left\|
\mathrm{R}(\mathcal{S}_{r,k,i})
-
\mathrm{R}(\widehat{\mathcal{S}}_i)
\right\|_1
}{
J\sigma_i
},
\quad
\sigma_i
=
\max\!\left(
\operatorname{std}(\widehat{\mathcal{S}}_i),10^{-12}
\right),
\end{equation}
where \(\mathrm{R}(\cdot)\) contains the \(J=9\) summary statistics
defined in Section~IV-B. The error is divided by the standard deviation
of the observed regime to reduce the effect of differences in scale
across regimes. The run-level mean calibration error is
\begin{equation}
\label{eq:EMCE}
E_{\mathrm{MCE}}^{(r)}
=
\frac{1}{K}
\sum_{k=1}^{K}
\sum_{i=1}^{C}
\omega_i e_{r,k,i},
\qquad
\omega_i
=
\frac{n_i}{\sum_{\ell=1}^{C}n_\ell},
\end{equation}
where \(n_i\) is the length of the \(i\)-th ground-truth regime.
A lower \(E_{\mathrm{MCE}}\) indicates better overall calibration.
The replay simulations are used only for post-hoc evaluation and are
excluded from the online simulator-evaluation budget.

\subsubsection{Convergence Performance}

The convergence performance \(P_{\mathrm{CON}}\) measures the average
anytime performance during online optimization. Let
\(F_{r,b,j}^{\mathrm{best}}\) denote the best-so-far fitness at iteration
\(j\) of the optimization performed for observation block \(b\), and let
\(\mathcal{V}_r\) denote all valid recorded iterations in run \(r\). We
define
\begin{equation}
\label{eq:PCON}
P_{\mathrm{CON}}^{(r)}
=
\frac{1}{|\mathcal{V}_r|}
\sum_{(b,j)\in\mathcal{V}_r}
F_{r,b,j}^{\mathrm{best}}.
\end{equation}
Optimization traces used only for change detection and subsequently
discarded are excluded from \(\mathcal{V}_r\). A lower
\(P_{\mathrm{CON}}\) indicates that the optimizer reaches and maintains
lower fitness values more rapidly.

\subsubsection{Number of Function Evaluations}

The number of function evaluations (NFE) counts all simulator
evaluations performed during online calibration, where each function
evaluation corresponds to one simulator execution. It includes
auxiliary detector evaluations and repeated optimization caused by
detected changes, but excludes the post-hoc replay simulations used to
compute \(E_{\mathrm{MCE}}\). A lower NFE indicates a lower online
calibration cost.

\subsubsection{Change Detection Metrics}

A detected change point is counted as a true positive (TP) only when it
exactly matches an unmatched ground-truth change point. An unmatched
detection is counted as a false positive (FP), whereas an unmatched
ground-truth change point is counted as a false negative (FN). We report
precision, recall, F1, and the average number of detected change points.
Higher precision, recall, and F1 indicate better detection performance,
while a detected count closer to the ground-truth count indicates less
over- or under-detection.

\subsubsection{Statistical Analysis}

Each method is independently executed 10 times on every benchmark
instance, and the mean and standard deviation are reported. Methods are
ranked within each instance according to their mean performance, and
average ranks are reported across instances. We additionally report
W/T/L counts and exact two-sided sign-test \(p\)-values between PosEDO
and each compared method, treating each instance as one paired sample.
A \(p\)-value below \(0.05\) is considered statistically
significant. Any transformations and per-instance best normalization
used for tabular presentation are specified in the corresponding table
captions.

\begin{table*}[htbp]
\centering
\caption{Main comparison on the Brock--Hommes benchmark. The table
reports best-normalized \(E_{\mathrm{MCE}}\), best-normalized
\(P_{\mathrm{CON}}\), and NFE. Before normalization,
\(E_{\mathrm{MCE}}\) is computed from scale-adjusted replay errors and
transformed using \(\log(1+E_{\mathrm{MCE}})\). A normalized value of
1 indicates the best method within an instance, and larger values are
worse. NFE is reported as mean\(\pm\)std over 10 independent runs.
DBD-Rand uses \(\alpha=0.4\), which achieves its highest average
detection F1 among the tested thresholds. The best result in each row
is highlighted in bold.}
\label{tab:bh-main-final9}
{\scriptsize
\begin{adjustbox}{max width=\textwidth,keepaspectratio}
\begin{tabular}{llcccccccccc}
\toprule
\textbf{Instance} & \textbf{Metric} & \textbf{PosEDO} & \textbf{FBCD-Arch} & \textbf{MFBCD-Arch} & \textbf{FBCD-DETOInit} & \textbf{MFBCD-DETOInit} & \textbf{FBCD-NNIT} & \textbf{MFBCD-NNIT} & \textbf{FBCD-Rand} & \textbf{MFBCD-Rand} & \textbf{DBD-Rand (\(\alpha=0.4\))}\\
\midrule
\multirow{3}{*}{\(F_1\)}
& \(E_{\mathrm{MCE}}\) & 1.022 & 1.096 & \textbf{1.000} & 1.031 & 1.004 & 1.047 & 1.283 & 1.016 & 1.039 & 1.394\\
& \(P_{\mathrm{CON}}\) & \textbf{1.000\(\pm\)0.021} & 1.518\(\pm\)0.069 & 1.252\(\pm\)0.046 & 1.530\(\pm\)0.085 & 1.266\(\pm\)0.044 & 1.495\(\pm\)0.060 & 1.250\(\pm\)0.073 & 1.521\(\pm\)0.067 & 1.229\(\pm\)0.096 & 2.014\(\pm\)0.264\\
& NFE & \textbf{15700.4\(\pm\)143.6} & 20003.0\(\pm\)653.2 & 17924.0\(\pm\)479.8 & 19920.4\(\pm\)829.4 & 17834.0\(\pm\)373.6 & 20195.6\(\pm\)778.4 & 17803.6\(\pm\)466.2 & 19796.8\(\pm\)481.4 & 17774.4\(\pm\)526.0 & 18486.0\(\pm\)189.6\\
\midrule
\multirow{3}{*}{\(F_2\)}
& \(E_{\mathrm{MCE}}\) & 1.020 & 1.248 & 1.026 & 1.165 & 1.025 & 1.148 & 1.061 & 1.271 & \textbf{1.000} & 1.638\\
& \(P_{\mathrm{CON}}\) & \textbf{1.000\(\pm\)0.026} & 1.476\(\pm\)0.087 & 1.181\(\pm\)0.087 & 1.482\(\pm\)0.087 & 1.124\(\pm\)0.081 & 1.412\(\pm\)0.072 & 1.126\(\pm\)0.041 & 1.503\(\pm\)0.068 & 1.209\(\pm\)0.056 & 1.776\(\pm\)0.281\\
& NFE & \textbf{15772.4\(\pm\)97.4} & 20867.2\(\pm\)839.6 & 17984.0\(\pm\)558.0 & 21113.4\(\pm\)772.6 & 17628.4\(\pm\)700.4 & 21603.2\(\pm\)1243.8 & 17557.8\(\pm\)371.8 & 21332.4\(\pm\)590.4 & 18098.8\(\pm\)541.6 & 18504.4\(\pm\)162.6\\
\midrule
\multirow{3}{*}{\(F_3\)}
& \(E_{\mathrm{MCE}}\) & 1.156 & 1.120 & 1.009 & 1.107 & 1.027 & \textbf{1.000} & 1.608 & 1.087 & 1.045 & 1.583\\
& \(P_{\mathrm{CON}}\) & \textbf{1.000\(\pm\)0.033} & 1.411\(\pm\)0.064 & 1.214\(\pm\)0.091 & 1.460\(\pm\)0.064 & 1.223\(\pm\)0.053 & 1.388\(\pm\)0.057 & 1.231\(\pm\)0.070 & 1.409\(\pm\)0.040 & 1.231\(\pm\)0.087 & 1.645\(\pm\)0.194\\
& NFE & \textbf{15681.0\(\pm\)96.8} & 21037.2\(\pm\)702.2 & 18471.8\(\pm\)559.0 & 21446.4\(\pm\)756.2 & 18770.0\(\pm\)547.2 & 21244.8\(\pm\)684.0 & 18746.6\(\pm\)673.0 & 20913.2\(\pm\)697.6 & 18465.6\(\pm\)643.8 & 18520.4\(\pm\)206.0\\
\midrule
\multirow{3}{*}{\(F_4\)}
& \(E_{\mathrm{MCE}}\) & \textbf{1.000} & 1.003 & 1.125 & 1.066 & 1.076 & 1.070 & 1.078 & 1.074 & 1.131 & 1.987\\
& \(P_{\mathrm{CON}}\) & \textbf{1.000\(\pm\)0.018} & 1.211\(\pm\)0.046 & 1.163\(\pm\)0.088 & 1.210\(\pm\)0.054 & 1.176\(\pm\)0.042 & 1.185\(\pm\)0.050 & 1.143\(\pm\)0.039 & 1.242\(\pm\)0.047 & 1.140\(\pm\)0.063 & 1.491\(\pm\)0.101\\
& NFE & \textbf{15756.8\(\pm\)135.2} & 20540.8\(\pm\)664.4 & 19006.0\(\pm\)580.2 & 20927.8\(\pm\)910.4 & 19097.2\(\pm\)453.6 & 20950.4\(\pm\)646.0 & 19036.0\(\pm\)415.0 & 20625.6\(\pm\)545.2 & 19061.2\(\pm\)475.2 & 18477.2\(\pm\)175.4\\
\midrule
\multirow{3}{*}{\(F_5\)}
& \(E_{\mathrm{MCE}}\) & \textbf{1.000} & 1.268 & 1.512 & 1.377 & 1.488 & 1.288 & 1.454 & 1.276 & 1.476 & 1.294\\
& \(P_{\mathrm{CON}}\) & \textbf{1.000\(\pm\)0.024} & 1.256\(\pm\)0.046 & 1.129\(\pm\)0.050 & 1.249\(\pm\)0.042 & 1.130\(\pm\)0.049 & 1.243\(\pm\)0.047 & 1.143\(\pm\)0.042 & 1.246\(\pm\)0.062 & 1.176\(\pm\)0.067 & 1.521\(\pm\)0.168\\
& NFE & \textbf{15847.6\(\pm\)100.0} & 21254.0\(\pm\)534.2 & 18462.8\(\pm\)410.4 & 20920.8\(\pm\)803.2 & 18585.2\(\pm\)502.8 & 21764.0\(\pm\)493.4 & 18711.6\(\pm\)485.4 & 21318.8\(\pm\)984.8 & 18795.2\(\pm\)930.2 & 18482.0\(\pm\)213.8\\
\midrule
\multirow{3}{*}{\(F_6\)}
& \(E_{\mathrm{MCE}}\) & \textbf{1.000} & 1.623 & 1.135 & 1.179 & 2.010 & 1.067 & 1.465 & 1.139 & 2.033 & 1.603\\
& \(P_{\mathrm{CON}}\) & \textbf{1.000\(\pm\)0.022} & 1.371\(\pm\)0.114 & 1.173\(\pm\)0.108 & 1.366\(\pm\)0.101 & 1.214\(\pm\)0.159 & 1.357\(\pm\)0.058 & 1.200\(\pm\)0.119 & 1.326\(\pm\)0.074 & 1.229\(\pm\)0.168 & 1.683\(\pm\)0.177\\
& NFE & \textbf{15849.2\(\pm\)109.6} & 20780.8\(\pm\)656.6 & 18528.8\(\pm\)552.4 & 20731.6\(\pm\)499.4 & 18406.8\(\pm\)516.6 & 21292.0\(\pm\)582.0 & 18725.8\(\pm\)562.4 & 20510.6\(\pm\)668.4 & 18383.2\(\pm\)513.4 & 18467.2\(\pm\)241.8\\
\midrule
\multirow{3}{*}{\(F_7\)}
& \(E_{\mathrm{MCE}}\) & \textbf{1.000} & 1.103 & 1.306 & 1.019 & 1.430 & 1.079 & 1.167 & 1.105 & 1.166 & 1.813\\
& \(P_{\mathrm{CON}}\) & \textbf{1.000\(\pm\)0.021} & 1.210\(\pm\)0.048 & 1.153\(\pm\)0.021 & 1.216\(\pm\)0.044 & 1.162\(\pm\)0.052 & 1.200\(\pm\)0.031 & 1.179\(\pm\)0.029 & 1.220\(\pm\)0.044 & 1.128\(\pm\)0.034 & 1.565\(\pm\)0.136\\
& NFE & \textbf{15894.6\(\pm\)102.8} & 21603.6\(\pm\)483.6 & 20719.2\(\pm\)307.4 & 21676.8\(\pm\)576.4 & 20394.0\(\pm\)343.6 & 21755.0\(\pm\)494.6 & 20607.8\(\pm\)222.8 & 21657.6\(\pm\)389.6 & 20670.0\(\pm\)397.4 & 18536.6\(\pm\)146.0\\
\midrule
\multirow{3}{*}{\(F_8\)}
& \(E_{\mathrm{MCE}}\) & 1.012 & 1.000 & 1.322 & \textbf{1.000} & 1.234 & 1.028 & 1.250 & 1.099 & 1.203 & 1.547\\
& \(P_{\mathrm{CON}}\) & \textbf{1.000\(\pm\)0.029} & 1.110\(\pm\)0.052 & 1.123\(\pm\)0.045 & 1.095\(\pm\)0.054 & 1.094\(\pm\)0.049 & 1.124\(\pm\)0.063 & 1.129\(\pm\)0.038 & 1.102\(\pm\)0.059 & 1.135\(\pm\)0.055 & 1.220\(\pm\)0.083\\
& NFE & \textbf{15959.2\(\pm\)115.0} & 22159.2\(\pm\)452.0 & 20249.6\(\pm\)341.8 & 22019.2\(\pm\)957.4 & 20296.0\(\pm\)276.8 & 21715.6\(\pm\)782.6 & 20439.6\(\pm\)316.4 & 21989.6\(\pm\)749.0 & 20210.2\(\pm\)374.2 & 18486.8\(\pm\)183.8\\
\midrule
\multirow{3}{*}{\(F_9\)}
& \(E_{\mathrm{MCE}}\) & 1.167 & 1.038 & 1.231 & 1.031 & 1.190 & \textbf{1.000} & 1.255 & 1.033 & 1.271 & 3.277\\
& \(P_{\mathrm{CON}}\) & \textbf{1.000\(\pm\)0.014} & 1.185\(\pm\)0.050 & 1.129\(\pm\)0.036 & 1.174\(\pm\)0.051 & 1.124\(\pm\)0.058 & 1.177\(\pm\)0.045 & 1.103\(\pm\)0.036 & 1.182\(\pm\)0.033 & 1.132\(\pm\)0.045 & 1.431\(\pm\)0.124\\
& NFE & \textbf{15929.2\(\pm\)158.2} & 22818.4\(\pm\)1004.2 & 19708.8\(\pm\)501.4 & 22152.4\(\pm\)771.6 & 19959.6\(\pm\)631.6 & 22630.8\(\pm\)876.0 & 19541.2\(\pm\)635.8 & 22679.8\(\pm\)587.0 & 19869.2\(\pm\)644.4 & 18672.4\(\pm\)231.6\\
\midrule
\multicolumn{12}{l}{\textit{Statistical summary based on \(E_{\mathrm{MCE}}\)}}\\
\multicolumn{2}{l}{Avg. rank\(\downarrow\)} & \textbf{2.89} & 5.00 & 5.78 & 4.11 & 6.00 & 3.56 & 7.44 & 4.67 & 6.56 & 9.00\\
\multicolumn{2}{l}{Sign-test \(p\) vs PosEDO} & -- & 0.508 & 0.180 & 0.508 & 0.180 & 0.180 & 0.004 & 0.508 & 0.180 & 0.004\\
\multicolumn{2}{l}{W/T/L vs PosEDO} & -- & 6/0/3 & 7/0/2 & 6/0/3 & 7/0/2 & 7/0/2 & 9/0/0 & 6/0/3 & 7/0/2 & 9/0/0\\
\midrule
\multicolumn{12}{l}{\textit{Statistical summary based on \(P_{\mathrm{CON}}\)}}\\
\multicolumn{2}{l}{Avg. rank\(\downarrow\)} & \textbf{1.00} & 7.67 & 3.44 & 7.33 & 3.44 & 6.33 & 3.89 & 7.44 & 4.44 & 10.00\\
\multicolumn{2}{l}{Sign-test \(p\) vs PosEDO} & -- & 0.004 & 0.004 & 0.004 & 0.004 & 0.004 & 0.004 & 0.004 & 0.004 & 0.004\\
\multicolumn{2}{l}{W/T/L vs PosEDO} & -- & 9/0/0 & 9/0/0 & 9/0/0 & 9/0/0 & 9/0/0 & 9/0/0 & 9/0/0 & 9/0/0 & 9/0/0\\
\midrule
\multicolumn{12}{l}{\textit{Statistical summary based on NFE}}\\
\multicolumn{2}{l}{Avg. rank\(\downarrow\)} & \textbf{1.00} & 8.33 & 4.00 & 8.33 & 4.44 & 9.33 & 4.22 & 8.00 & 3.89 & 3.44\\
\multicolumn{2}{l}{Sign-test \(p\) vs PosEDO} & -- & 0.004 & 0.004 & 0.004 & 0.004 & 0.004 & 0.004 & 0.004 & 0.004 & 0.004\\
\multicolumn{2}{l}{W/T/L vs PosEDO} & -- & 9/0/0 & 9/0/0 & 9/0/0 & 9/0/0 & 9/0/0 & 9/0/0 & 9/0/0 & 9/0/0 & 9/0/0\\
\bottomrule
\end{tabular}
\end{adjustbox}
}
\end{table*}

\subsection{Experimental Configuration}

Unless otherwise specified, all methods use the same AMP/PSO
configuration, with a maximum of \(I_{\max}=10\) optimization iterations
and a fixed total population size of \(\lambda=40\). The same parameter
bounds and optimizer settings are used across all compared methods.

The posterior estimator \(q_\phi\) is randomly initialized and trained
online, in parallel with evolutionary optimization, using the
parameter--trajectory pairs generated during simulator evaluation.
Online posterior training uses a batch size of \(B=200\) and a learning
rate of \(\eta=5\times10^{-4}\), and is terminated when the training loss
does not improve for three consecutive epochs.

For posterior-based change detection and population adaptation,
\(N=2000\) posterior samples are used. The KL-divergence threshold
\(\varepsilon\) is set to \(30\) for BH, \(0.05\) for PGPS-6D, and
\(0.2\) for the higher-dimensional PGPS benchmarks. Each threshold is
fixed across all instances within the corresponding benchmark. The
sensitivity of the BH threshold is examined in
Section~\ref{sec:bh-threshold-sensitivity}. For DBD-Rand, we evaluate
\(\alpha\in\{0,0.2,0.4,0.6,0.8,1.0\}\).

All benchmark instances are generated in advance and shared by all
methods. Each method is independently executed 10 times per instance
using different random seeds. To compute \(E_{\mathrm{MCE}}\), the
calibrated parameters from each run are independently replayed \(K=5\)
times. All simulator calls performed during online calibration are
included in NFE, whereas replay simulations used only for post-hoc
evaluation are excluded. The mean and standard deviation over the 10
independent runs are reported.

\section{Experimental Results on the BH Model}
\label{sec:results1}

The BH experiments evaluate PosEDO in terms of overall calibration
accuracy, convergence performance, simulator-evaluation cost,
change-detection reliability, threshold sensitivity, and
posterior-guided adaptation.

\subsection{Overall Performance and Evaluation Cost}

Table~\ref{tab:bh-main-final9} compares PosEDO with the EDO baselines
on the nine BH instances. For DBD-Rand, the table reports
\(\alpha=0.4\), which achieves its highest average detection F1 among
the tested thresholds.
PosEDO requires the fewest simulator evaluations on all nine instances,
with an average NFE rank of \(1.00\), a W/T/L result of \(9/0/0\), and
a sign-test \(p\)-value of \(0.004\) against every baseline. The higher
costs of the FBCD- and MFBCD-based methods result from repeated
optimization after detected changes, while DBD-Rand requires additional
detector re-evaluations. In contrast, posterior training, change
detection, and population adaptation in PosEDO introduce no additional
simulator evaluations.

PosEDO also achieves the best average ranks of \(2.89\) and \(1.00\)
for \(E_{\mathrm{MCE}}\) and \(P_{\mathrm{CON}}\), respectively. It
obtains the lowest \(E_{\mathrm{MCE}}\) on four instances and
outperforms each baseline on six to nine instances. For
\(P_{\mathrm{CON}}\), PosEDO ranks first on all nine instances and is
significantly better than every baseline. Thus, its lower
simulator-evaluation cost does not compromise calibration accuracy or
online convergence.

Because the overall results reflect the combined effects of change
detection and population adaptation, the following subsections examine
the two components separately.

\begin{figure*}[t]
    \centering
    \includegraphics[width=0.95\textwidth]
    {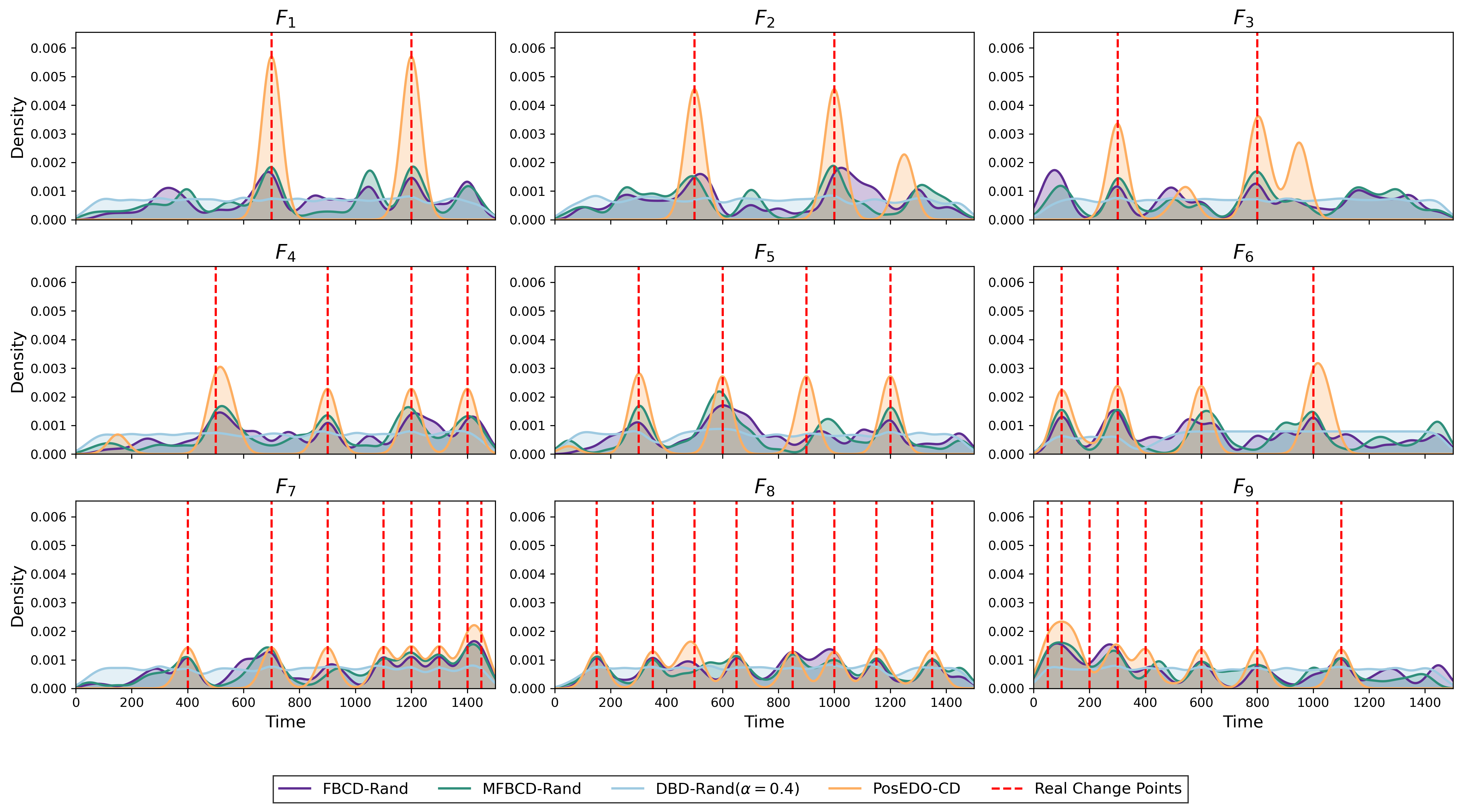}
    \caption{Temporal distributions of detected change points on the
    nine Brock--Hommes instances. All methods use random population
    re-initialization with the same random seeds. Red dashed lines indicate ground-truth change
    points. PosEDO-CD produces detections concentrated around the true
    regime boundaries, whereas the other methods exhibit more dispersed
    detections and false alarms.}
    \label{fig:bh-detection-density}
\end{figure*}

\subsection{Effect of Posterior-based Change Detection}

We next examine whether posterior shifts provide a reliable signal for
identifying regime changes. PosEDO and PosEDO-CD both use
posterior-based detection, while PosEDO-CD replaces posterior-guided
adaptation with random re-initialization. The latter is therefore used
for controlled comparison with the other Rand-based methods.

Table~\ref{tab:bh-change-detection-overall-all-alpha-final9} reports the
results under the strict exact-match criterion. PosEDO achieves the
highest precision and F1 score, reaching \(0.955\) and \(0.954\),
respectively, with a recall of \(0.957\). PosEDO-CD obtains similar
results, with a precision of \(0.946\), a recall of \(0.958\), and an
F1 score of \(0.949\). Their average detected counts, \(4.600\) and
\(4.644\), are both close to the average of \(4.67\) ground-truth
changes.

The FBCD-based methods retain high recall but detect approximately ten
changes on average, resulting in precision values around \(0.41\) and
F1 scores around \(0.55\). MFBCD reduces over-detection, but its F1
scores remain around \(0.70\). DBD-Rand is sensitive to \(\alpha\):
small thresholds produce excessive detections, whereas large thresholds
lead to substantial missed changes. Its best tested F1 score remains
considerably lower than those of PosEDO and PosEDO-CD.

Fig.~\ref{fig:bh-detection-density} further visualizes the temporal
distribution of detected changes. The detections produced by PosEDO-CD
are concentrated around the ground-truth regime boundaries, whereas the
other methods produce more dispersed detections at ordinary observation
updates. This pattern is consistent with the higher precision and F1
reported in Table~\ref{tab:bh-change-detection-overall-all-alpha-final9}.

\begin{figure}[t]
    \centering
    \includegraphics[width=\columnwidth]
    {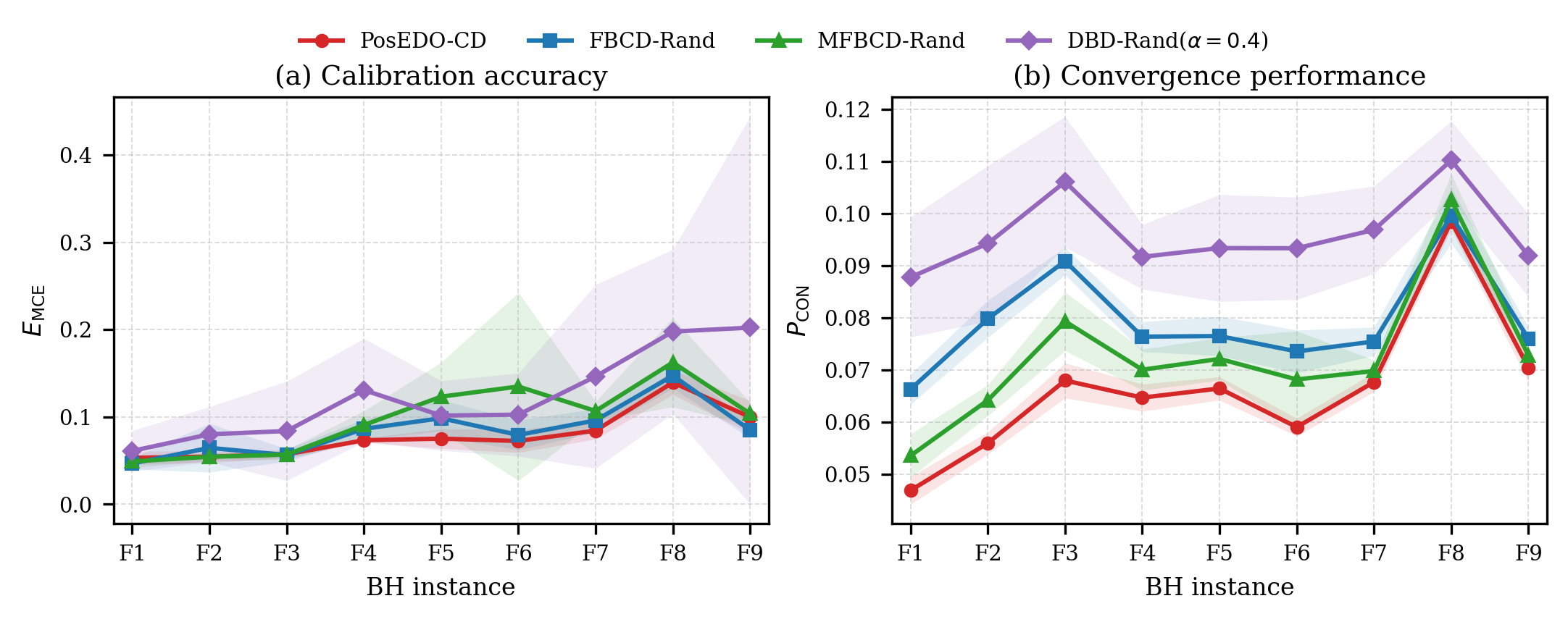}
    \caption{Controlled comparison of change-detection mechanisms on
    the Brock--Hommes benchmark. All methods use random population
    re-initialization after a detected change. (a) Scale-adjusted
    \(E_{\mathrm{MCE}}\). (b) \(P_{\mathrm{CON}}\). Lower values are
    better, and shaded areas indicate mean\(\pm\)standard deviation
    over 10 runs.}
    \label{fig:bh-detection-controlled-comparison}
\end{figure}

To examine the downstream effect of detection quality,
Fig.~\ref{fig:bh-detection-controlled-comparison} compares PosEDO-CD,
FBCD-Rand, MFBCD-Rand, and DBD-Rand(\(\alpha=0.4\)). Since all four
methods use random re-initialization after a detected change, their
differences primarily reflect their detection mechanisms. PosEDO-CD
achieves lower \(E_{\mathrm{MCE}}\) and \(P_{\mathrm{CON}}\) on most
instances, showing that more reliable regime-boundary detection improves
both calibration accuracy and subsequent online optimization.

\begin{table}[htbp]
\centering
\caption{Overall change-detection performance of all compared methods on the Brock--Hommes benchmark. Precision, recall, F1, and the average number of detected change points (Detected) are averaged over the nine instances and 10 independent runs per instance. A detected change point is counted as correct only when it exactly matches a ground-truth change point, corresponding to a tolerance window of zero. Higher precision, recall, and F1 values are better. The nine instances contain an average of \(4.67\) ground-truth change points, and hence a Detected value closer to \(4.67\) is preferred. DBD-Rand results are reported for all tested \(\alpha\) values.}
\label{tab:bh-change-detection-overall-all-alpha-final9}
{\scriptsize
\begin{adjustbox}{max width=0.5\textwidth, keepaspectratio}
\begin{tabular}{lcccc}
\toprule
\textbf{Method} & \textbf{Precision} & \textbf{Recall} & \textbf{F1} & \textbf{Detected}\\
\midrule
PosEDO & \textbf{0.955$\pm$0.098} & 0.957$\pm$0.094
& \textbf{0.954$\pm$0.089} & 4.600$\pm$2.341\\
PosEDO-CD & 0.946$\pm$0.103 & 0.958$\pm$0.092
& 0.949$\pm$0.086 & \textbf{4.644$\pm$2.329}\\
FBCD-Arch & 0.411$\pm$0.190 & 0.928$\pm$0.117 & 0.545$\pm$0.185 & 10.156$\pm$1.841\\
MFBCD-Arch & 0.618$\pm$0.179 & 0.876$\pm$0.136 & 0.705$\pm$0.132 & 6.156$\pm$1.937\\
FBCD-DETOInit & 0.429$\pm$0.215 & 0.944$\pm$0.114 & 0.562$\pm$0.207 & 10.111$\pm$1.751\\
MFBCD-DETOInit & 0.617$\pm$0.187 & 0.860$\pm$0.135 & 0.698$\pm$0.138 & 6.111$\pm$1.968\\
FBCD-NNIT & 0.415$\pm$0.210 & 0.953$\pm$0.105 & 0.549$\pm$0.201 & 10.544$\pm$1.717\\
MFBCD-NNIT & 0.619$\pm$0.183 & 0.867$\pm$0.140 & 0.703$\pm$0.141 & 6.144$\pm$1.975\\
FBCD-Rand & 0.416$\pm$0.201 & 0.928$\pm$0.120 & 0.549$\pm$0.196 & 10.133$\pm$1.794\\
MFBCD-Rand & 0.609$\pm$0.192 & 0.868$\pm$0.147 & 0.695$\pm$0.144 & 6.211$\pm$1.957\\
DBD-Rand($\alpha=0$) & 0.161$\pm$0.086 & \textbf{1.000$\pm$0.000} & 0.268$\pm$0.126 & 29.000$\pm$0.000\\
DBD-Rand($\alpha=0.2$) & 0.173$\pm$0.106 & 0.969$\pm$0.116 & 0.281$\pm$0.144 & 26.978$\pm$4.424\\
DBD-Rand($\alpha=0.4$) & 0.213$\pm$0.131 & 0.856$\pm$0.241 & 0.316$\pm$0.158 & 20.400$\pm$8.657\\
DBD-Rand($\alpha=0.6$) & 0.220$\pm$0.144 & 0.667$\pm$0.336 & 0.304$\pm$0.173 & 14.167$\pm$7.703\\
DBD-Rand($\alpha=0.8$) & 0.192$\pm$0.151 & 0.369$\pm$0.258 & 0.235$\pm$0.169 & 8.833$\pm$2.584\\
DBD-Rand($\alpha=1.0$) & 0.152$\pm$0.149 & 0.243$\pm$0.225 & 0.173$\pm$0.157 & 6.922$\pm$1.317\\
\bottomrule
\end{tabular}
\end{adjustbox}
}
\end{table}

\subsection{Sensitivity to the Detection Threshold}
\label{sec:bh-threshold-sensitivity}

\begin{figure}[t]
    \centering
    \includegraphics[width=0.5\textwidth]
    {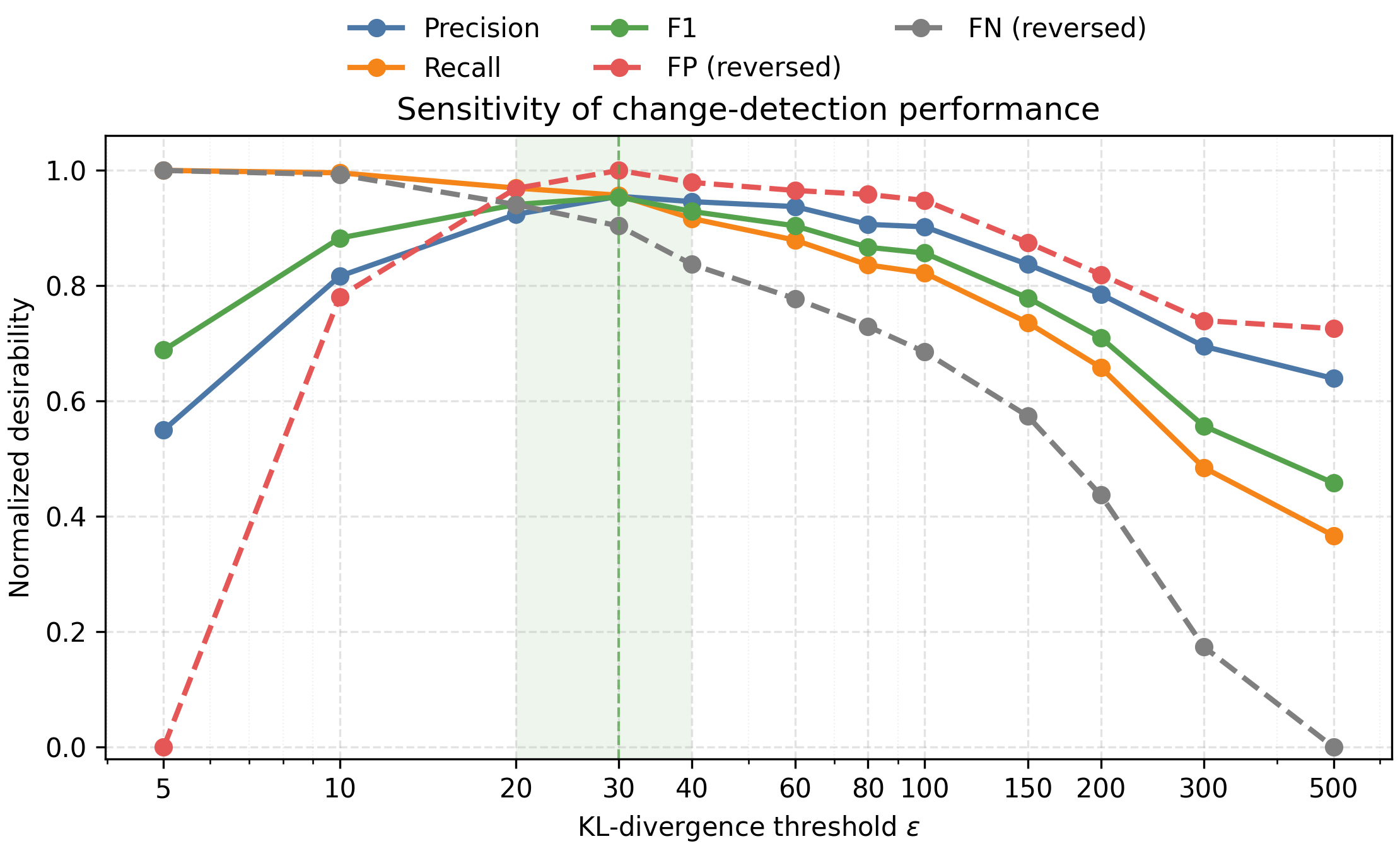}
    \caption{Sensitivity of posterior-based change detection to the
    KL-divergence threshold \(\varepsilon\) on the Brock--Hommes
    benchmark. Precision, recall, and F1 are shown together with the
    normalized desirability scores of false positives (FP) and false
    negatives (FN). For FP and FN, the values are reversely normalized,
    so larger values indicate fewer detection errors. The dashed
    vertical line marks the selected threshold \(\varepsilon=30\), and
    the shaded region highlights the favorable range
    \(\varepsilon\in[20,40]\).}
    \label{fig:bh-epsilon-sensitivity}
\end{figure}

Fig.~\ref{fig:bh-epsilon-sensitivity} shows the sensitivity of
posterior-based change detection to the KL-divergence threshold
\(\varepsilon\) on the BH benchmark. When \(\varepsilon\) is too small,
the detector becomes overly sensitive. In this case, recall remains high,
but precision is low because small posterior fluctuations can easily
trigger false alarms. As \(\varepsilon\) increases, false positives are
substantially reduced and precision improves accordingly.

However, when \(\varepsilon\) becomes too large, the detector turns
overly conservative. Recall and the desirability score associated with
false negatives both decrease, indicating that more true regime changes
are missed. Overall, the F1 score remains high and relatively stable for
\(\varepsilon \in [20,40]\), and reaches its best value around
\(\varepsilon=30\). This threshold also provides a good balance between
precision and recall. We therefore use \(\varepsilon=30\) as the default
setting in the BH experiments.

\subsection{Effect of Posterior-guided Adaptation}
\label{sec:adaptation}

To isolate the effect of population adaptation, all methods are provided
with the ground-truth change points. They therefore optimize over the
same calibration windows and differ only in how the population is
initialized after a regime change. We compare posterior-guided
adaptation, denoted as PosEDO-Ada, with random re-initialization (Rand),
archive-based adaptation (Arch), neural-network-based information
transfer (NNIT), and DETO-based initialization (DETOInit).

As shown in
Table~\ref{tab:bh-oracle-adaptation-normalized-f6-f8-f9-replay30},
PosEDO-Ada achieves the best average \(E_{\mathrm{MCE}}\) rank of
\(2.00\). It obtains the lowest calibration error on five of the nine
instances and outperforms Rand, Arch, NNIT, and DETOInit on seven,
eight, six, and six instances, respectively. These results indicate that
posterior samples generally provide useful candidate parameters after a
regime change, although the calibration improvement varies across
instances.

The advantage is more consistent for convergence. PosEDO-Ada obtains
the lowest \(P_{\mathrm{CON}}\) on all nine instances, with an average
rank of \(1.00\), a W/T/L result of \(9/0/0\), and a sign-test
\(p\)-value of \(0.004\) against every compared strategy. Since all
methods use the same ground-truth change points, these differences arise
from their population-initialization mechanisms.

\begin{figure*}[t]
    \centering
    \includegraphics[width=0.95\textwidth]
    {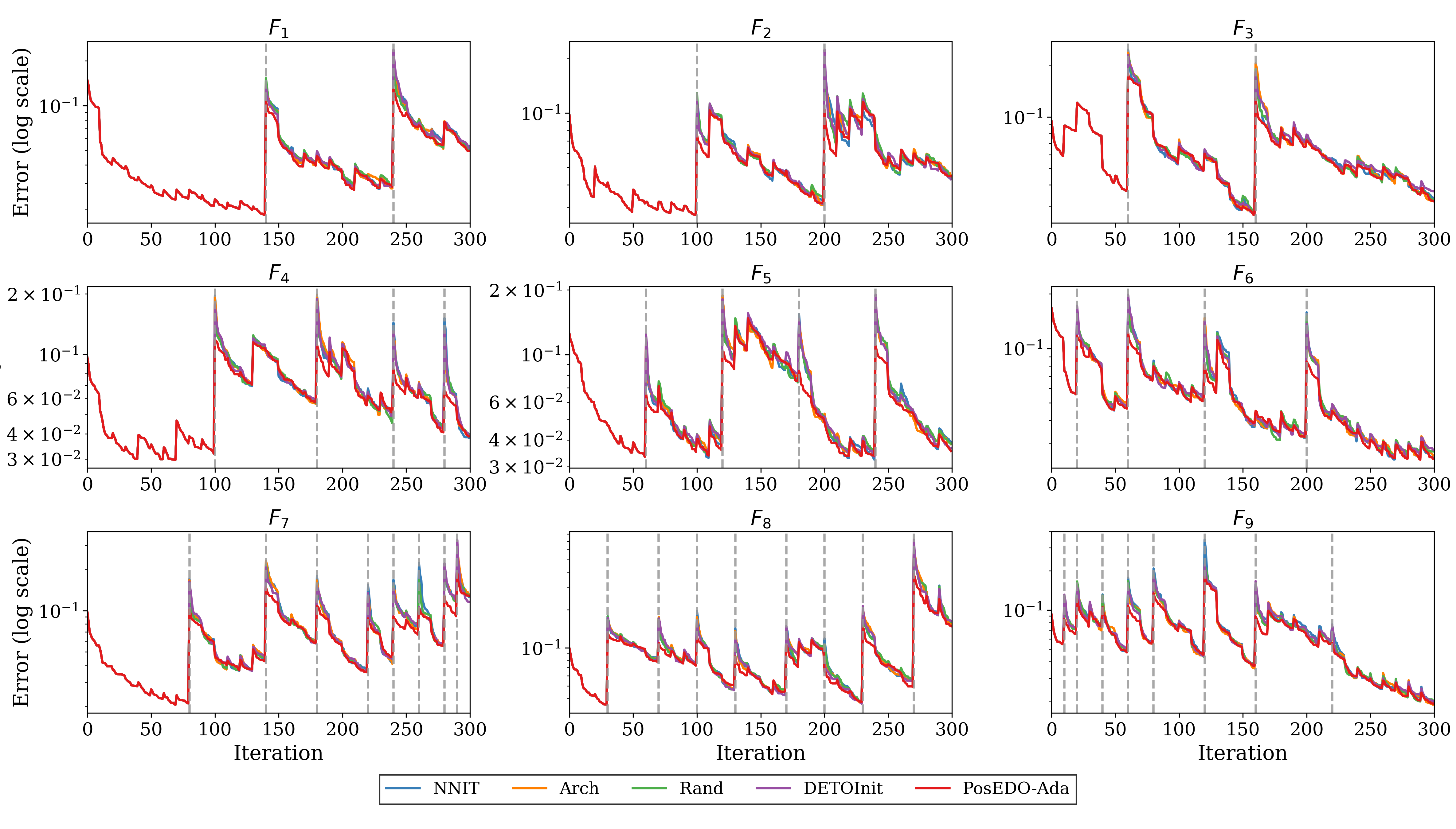}
    \caption{Comparisons of Best-fitness trajectories of population-adaptation
    strategies on the Brock--Hommes benchmark with known ground-truth change
    points. Lower values are better, and the vertical dashed lines
    indicate real regime changes. PosEDO-Ada uses posterior-guided
    adaptation; the other methods use Rand, Arch, NNIT, or DETOInit. The trajectories of different strategies remain the same within the first regime because they use the same backbone evolutionary search method (AMP/PSO) and the same random seeds.}
    \label{fig:bh-oracle-adaptation-convergence}
\end{figure*}

Fig.~\ref{fig:bh-oracle-adaptation-convergence} further shows the
best-fitness trajectories. Fitness increases when a new regime begins,
as indicated by the vertical dashed lines. PosEDO-Ada generally returns
to lower fitness more rapidly after these changes, consistent with its
superior \(P_{\mathrm{CON}}\). Overall, the observation-conditioned
posterior provides informative initial candidates for the new regime,
with its clearest benefit observed in post-change convergence.

\begin{table}[htbp]
\centering
\caption{Comparison of population-adaptation strategies on the
Brock--Hommes benchmark under ground-truth change points. The table
reports best-normalized \(E_{\mathrm{MCE}}\), computed from
scale-adjusted replay errors, and best-normalized
\(P_{\mathrm{CON}}\). A value of 1 indicates the best method within an
instance, and larger values are worse. Results are averaged over 10
independent runs, and W/T/L counts are reported from the perspective of
PosEDO-Ada.}
\label{tab:bh-oracle-adaptation-normalized-f6-f8-f9-replay30}
{\scriptsize
\begin{adjustbox}{max width=0.5\textwidth,keepaspectratio}
\begin{tabular}{llccccc}
\toprule
\textbf{Instance} & \textbf{Metric} & \textbf{Rand} & \textbf{Arch} & \textbf{NNIT} & \textbf{DETOInit} & \textbf{PosEDO-Ada}\\
\midrule
\multirow{2}{*}{\(F_1\)} & \(E_{\mathrm{MCE}}\) & 1.117 & 1.150 & 1.158 & 1.158 & \textbf{1.000}\\
& \(P_{\mathrm{CON}}\) & 1.050\(\pm\)0.023 & 1.050\(\pm\)0.023 & 1.057\(\pm\)0.027 & 1.064\(\pm\)0.020 & \textbf{1.000\(\pm\)0.016}\\
\midrule
\multirow{2}{*}{\(F_2\)} & \(E_{\mathrm{MCE}}\) & 1.025 & 1.011 & 1.042 & 1.025 & \textbf{1.000}\\
& \(P_{\mathrm{CON}}\) & 1.061\(\pm\)0.040 & 1.051\(\pm\)0.049 & 1.040\(\pm\)0.021 & 1.051\(\pm\)0.019 & \textbf{1.000\(\pm\)0.015}\\
\midrule
\multirow{2}{*}{\(F_3\)} & \(E_{\mathrm{MCE}}\) & 1.046 & 1.069 & 1.021 & 1.047 & \textbf{1.000}\\
& \(P_{\mathrm{CON}}\) & 1.038\(\pm\)0.022 & 1.041\(\pm\)0.019 & 1.027\(\pm\)0.022 & 1.047\(\pm\)0.024 & \textbf{1.000\(\pm\)0.020}\\
\midrule
\multirow{2}{*}{\(F_4\)} & \(E_{\mathrm{MCE}}\) & 1.024 & 1.051 & \textbf{1.000} & 1.011 & 1.045\\
& \(P_{\mathrm{CON}}\) & 1.047\(\pm\)0.026 & 1.058\(\pm\)0.018 & 1.056\(\pm\)0.018 & 1.056\(\pm\)0.026 & \textbf{1.000\(\pm\)0.018}\\
\midrule
\multirow{2}{*}{\(F_5\)} & \(E_{\mathrm{MCE}}\) & 1.026 & 1.041 & 1.009 & \textbf{1.000} & 1.039\\
& \(P_{\mathrm{CON}}\) & 1.072\(\pm\)0.021 & 1.068\(\pm\)0.023 & 1.065\(\pm\)0.029 & 1.080\(\pm\)0.037 & \textbf{1.000\(\pm\)0.020}\\
\midrule
\multirow{2}{*}{\(F_6\)} & \(E_{\mathrm{MCE}}\) & 1.013 & 1.031 & 1.036 & 1.020 & \textbf{1.000}\\
& \(P_{\mathrm{CON}}\) & 1.065\(\pm\)0.029 & 1.062\(\pm\)0.015 & 1.065\(\pm\)0.017 & 1.069\(\pm\)0.024 & \textbf{1.000\(\pm\)0.021}\\
\midrule
\multirow{2}{*}{\(F_7\)} & \(E_{\mathrm{MCE}}\) & 1.061 & \textbf{1.000} & 1.133 & 1.081 & 1.057\\
& \(P_{\mathrm{CON}}\) & 1.097\(\pm\)0.029 & 1.113\(\pm\)0.026 & 1.139\(\pm\)0.028 & 1.105\(\pm\)0.029 & \textbf{1.000\(\pm\)0.019}\\
\midrule
\multirow{2}{*}{\(F_8\)} & \(E_{\mathrm{MCE}}\) & 1.024 & 1.071 & 1.009 & \textbf{1.000} & 1.011\\
& \(P_{\mathrm{CON}}\) & 1.109\(\pm\)0.021 & 1.111\(\pm\)0.028 & 1.100\(\pm\)0.029 & 1.095\(\pm\)0.029 & \textbf{1.000\(\pm\)0.012}\\
\midrule
\multirow{2}{*}{\(F_9\)} & \(E_{\mathrm{MCE}}\) & 1.006 & 1.007 & 1.010 & 1.018 & \textbf{1.000}\\
& \(P_{\mathrm{CON}}\) & 1.073\(\pm\)0.021 & 1.065\(\pm\)0.027 & 1.115\(\pm\)0.032 & 1.080\(\pm\)0.030 & \textbf{1.000\(\pm\)0.019}\\
\midrule
\multicolumn{7}{l}{\textit{Statistical summary based on \(E_{\mathrm{MCE}}\)}}\\
\multicolumn{2}{l}{Avg. rank\(\downarrow\)} & 2.83 & 3.67 & 3.39 & 3.11 & \textbf{2.00}\\
\multicolumn{2}{l}{Sign-test \(p\) vs PosEDO-Ada} & 0.180 & 0.039 & 0.508 & 0.508 & --\\
\multicolumn{2}{l}{W/T/L vs PosEDO-Ada} & 7/0/2 & 8/0/1 & 6/0/3 & 6/0/3 & --\\
\midrule
\multicolumn{7}{l}{\textit{Statistical summary based on \(P_{\mathrm{CON}}\)}}\\
\multicolumn{2}{l}{Avg. rank\(\downarrow\)} & 3.22 & 3.44 & 3.33 & 4.00 & \textbf{1.00}\\
\multicolumn{2}{l}{Sign-test \(p\) vs PosEDO-Ada} & 0.004 & 0.004 & 0.004 & 0.004 & --\\
\multicolumn{2}{l}{W/T/L vs PosEDO-Ada} & 9/0/0 & 9/0/0 & 9/0/0 & 9/0/0 & --\\
\bottomrule
\end{tabular}
\end{adjustbox}
}
\end{table}

Overall, the BH experiments show that PosEDO improves calibration,
convergence, and simulator-evaluation efficiency. The controlled
experiments further verify the contributions of posterior-based change
detection and posterior-guided adaptation. We next examine whether these
advantages persist on the more stochastic PGPS simulator and its
higher-dimensional variants.

\section{Experimental Results on the PGPS Model}
\label{sec:results2}

The PGPS model provides a more stochastic and interaction-intensive
calibration problem than the BH model. We examine whether PosEDO remains
effective on the six-dimensional benchmark and its higher-dimensional
extensions.

\subsection{Overall Performance and Change Detection on the
Six-dimensional PGPS Model}

\begin{figure}[t]
    \centering
    \includegraphics[width=0.5\textwidth]
    {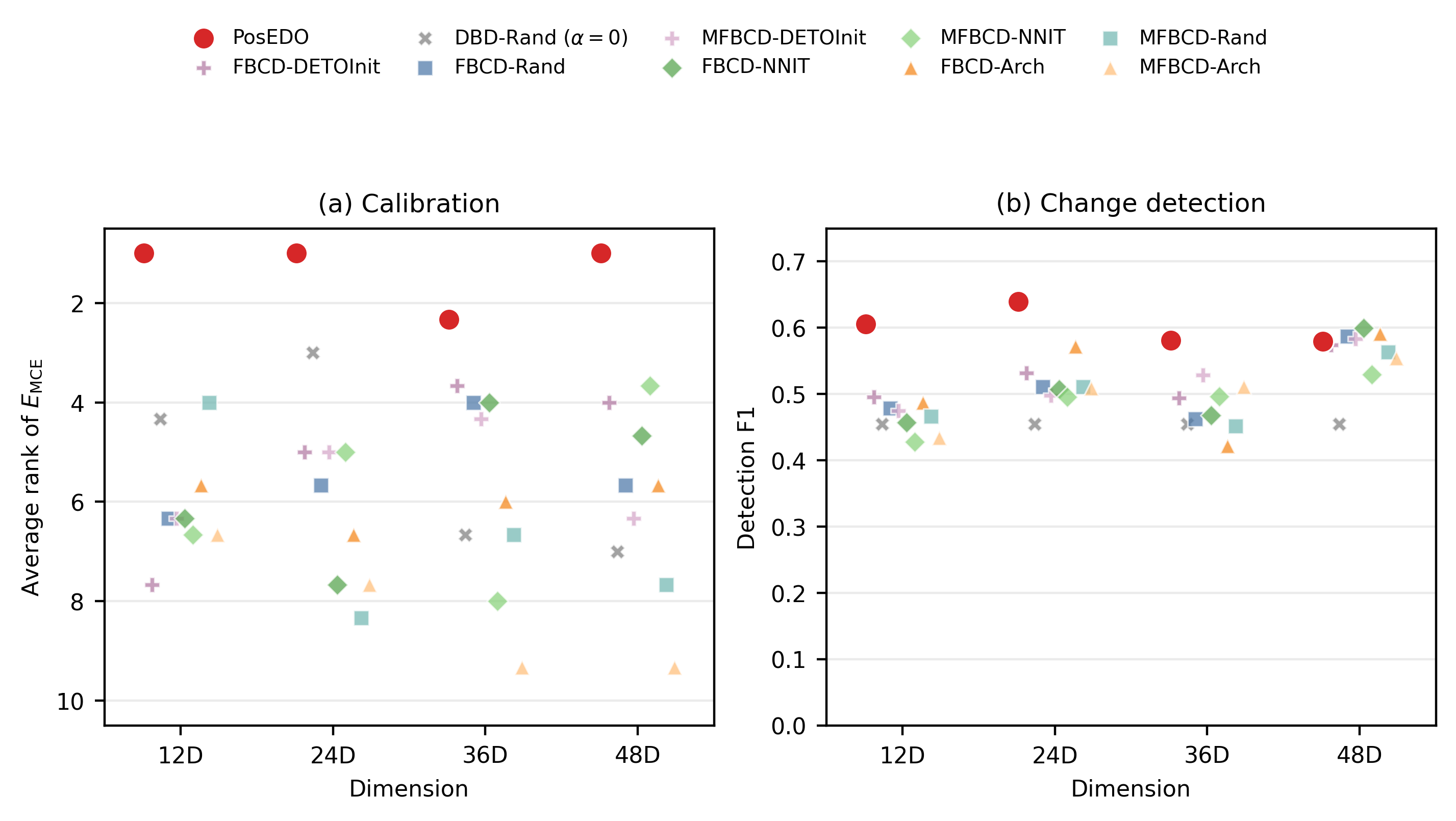}
    \caption{Performance comparison on the higher-dimensional PGPS
    benchmarks. (a) Average rank based on \(E_{\mathrm{MCE}}\), where
    a smaller value indicates better calibration performance.
    (b) Change-detection F1 score, where a larger value indicates better
    detection performance. Results are reported for
    \(d\in\{12,24,36,48\}\).}
    \label{fig:pgps-high-dimensional}
\end{figure}

Tables~\ref{tab:pgps-main} and
\ref{tab:pgps-change-detection-overall-all-alpha} report the overall
performance and change-detection results on the PGPS-6D benchmark.

PosEDO achieves the best average \(E_{\mathrm{MCE}}\) rank of \(2.222\),
obtains the lowest value on six of the nine instances, and outperforms
each baseline on seven or eight instances. It also requires the fewest
simulator evaluations on all nine instances, with an average NFE rank of
\(1.000\), a W/T/L result of \(9/0/0\), and a sign-test \(p\)-value of
\(0.004\) against every baseline. Thus, the calibration-accuracy and
evaluation-efficiency advantages observed on BH persist in the more
stochastic PGPS environment.

In contrast, PosEDO obtains an average \(P_{\mathrm{CON}}\) rank of
\(6.389\) and is less competitive than several baselines. This difference
is not contradictory: \(E_{\mathrm{MCE}}\) evaluates the replay accuracy
of the calibrated regime parameters, whereas \(P_{\mathrm{CON}}\)
summarizes the best-so-far fitness trajectories during online
optimization. The main advantage of PosEDO on PGPS therefore lies in
regime-level calibration accuracy and simulator-evaluation efficiency
rather than average convergence performance.

Change detection also becomes more difficult than on BH. PosEDO obtains
a precision of \(0.822\), a recall of \(0.816\), and the highest F1 score
of \(0.819\), while detecting \(4.933\) changes on average, close to the
five ground-truth changes. The FBCD- and MFBCD-based methods generally
detect \(6.6\)--\(7.5\) changes and exhibit lower precision, whereas
DBD-Rand is highly sensitive to \(\alpha\). Overall, posterior-based
detection retains a better balance between identifying true changes and
avoiding false detections, although its accuracy decreases under the
stronger stochasticity of PGPS.

\begin{table*}[htbp]
\centering
\caption{Main comparison on the PGPS-6D benchmark in terms of calibration accuracy, convergence performance, and online simulator evaluations. The table reports best-normalized \(E_{\mathrm{MCE}}\), best-normalized \(P_{\mathrm{CON}}\), and NFE. Before normalization, the source \(E_{\mathrm{MCE}}\) and \(P_{\mathrm{CON}}\) values are transformed using \(\log(1+x)\). For the normalized metrics, 1 indicates the best method within each instance row, and larger values indicate worse performance. NFE is reported as mean\(\pm\)std over 10 independent runs. For DBD-Rand, \(\alpha=0\) is reported because it achieves the highest change-detection F1 score among the tested values \(\alpha\in\{0,0.2,0.4,0.6,0.8,1.0\}\).}
\label{tab:pgps-main}
{\scriptsize
\begin{adjustbox}{max width=\textwidth,keepaspectratio}
\begin{tabular}{llcccccccccc}
\toprule
\textbf{Instance}
& \textbf{Metric}
& \textbf{PosEDO}
& \textbf{FBCD-Arch}
& \textbf{MFBCD-Arch}
& \textbf{FBCD-DETOInit}
& \textbf{MFBCD-DETOInit}
& \textbf{FBCD-NNIT}
& \textbf{MFBCD-NNIT}
& \textbf{FBCD-Rand}
& \textbf{MFBCD-Rand}
& \textbf{DBD-Rand (\(\alpha=0\))}\\
\midrule

\multirow{3}{*}{\(G_1\)}
& \(E_{\mathrm{MCE}}\)
& \textbf{1.000\(\pm\)0.044}
& 1.055\(\pm\)0.023
& 1.179\(\pm\)0.242
& 1.106\(\pm\)0.038
& 1.268\(\pm\)0.328
& 1.151\(\pm\)0.225
& 1.118\(\pm\)0.066
& 1.095\(\pm\)0.039
& 1.166\(\pm\)0.095
& 1.122\(\pm\)0.008\\
& \(P_{\mathrm{CON}}\)
& 1.039\(\pm\)0.041
& 1.037\(\pm\)0.049
& 1.050\(\pm\)0.056
& 1.026\(\pm\)0.023
& 1.086\(\pm\)0.078
& \textbf{1.000\(\pm\)0.036}
& 1.043\(\pm\)0.058
& 1.019\(\pm\)0.034
& 1.038\(\pm\)0.031
& 1.167\(\pm\)0.084\\
& NFE
& \textbf{9029\(\pm\)105}
& 10912\(\pm\)454
& 11088\(\pm\)798
& 10868\(\pm\)417
& 11308\(\pm\)952
& 10692\(\pm\)417
& 11264\(\pm\)694
& 11000\(\pm\)508
& 11484\(\pm\)732
& 11171\(\pm\)54\\
\midrule

\multirow{3}{*}{\(G_2\)}
& \(E_{\mathrm{MCE}}\)
& 1.001\(\pm\)0.103
& 1.044\(\pm\)0.099
& \textbf{1.000\(\pm\)0.025}
& 1.003\(\pm\)0.018
& 1.007\(\pm\)0.029
& 1.015\(\pm\)0.008
& 1.024\(\pm\)0.064
& 1.011\(\pm\)0.013
& 1.007\(\pm\)0.021
& 1.021\(\pm\)0.009\\
& \(P_{\mathrm{CON}}\)
& 1.088\(\pm\)0.057
& 1.003\(\pm\)0.027
& 1.046\(\pm\)0.069
& 1.019\(\pm\)0.040
& 1.003\(\pm\)0.061
& 1.007\(\pm\)0.049
& 1.039\(\pm\)0.059
& \textbf{1.000\(\pm\)0.061}
& 1.004\(\pm\)0.023
& 1.147\(\pm\)0.151\\
& NFE
& \textbf{9087\(\pm\)178}
& 10516\(\pm\)603
& 11308\(\pm\)777
& 10472\(\pm\)580
& 10912\(\pm\)541
& 10296\(\pm\)371
& 11352\(\pm\)404
& 10604\(\pm\)438
& 11178\(\pm\)512
& 11162\(\pm\)71\\
\midrule

\multirow{3}{*}{\(G_3\)}
& \(E_{\mathrm{MCE}}\)
& \textbf{1.000\(\pm\)0.058}
& 1.091\(\pm\)0.032
& 1.092\(\pm\)0.031
& 1.050\(\pm\)0.066
& 1.107\(\pm\)0.026
& 1.090\(\pm\)0.013
& 1.146\(\pm\)0.122
& 1.090\(\pm\)0.015
& 1.099\(\pm\)0.022
& 1.108\(\pm\)0.002\\
& \(P_{\mathrm{CON}}\)
& 1.052\(\pm\)0.040
& 1.028\(\pm\)0.048
& 1.036\(\pm\)0.052
& \textbf{1.000\(\pm\)0.043}
& 1.069\(\pm\)0.086
& 1.028\(\pm\)0.044
& 1.084\(\pm\)0.091
& 1.045\(\pm\)0.037
& 1.118\(\pm\)0.115
& 1.296\(\pm\)0.158\\
& NFE
& \textbf{8946\(\pm\)138}
& 10604\(\pm\)702
& 10912\(\pm\)743
& 11044\(\pm\)527
& 10737\(\pm\)695
& 10912\(\pm\)541
& 11044\(\pm\)438
& 10692\(\pm\)510
& 10780\(\pm\)475
& 11176\(\pm\)38\\
\midrule

\multirow{3}{*}{\(G_4\)}
& \(E_{\mathrm{MCE}}\)
& \textbf{1.000\(\pm\)0.043}
& 1.061\(\pm\)0.033
& 1.186\(\pm\)0.133
& 1.083\(\pm\)0.048
& 1.192\(\pm\)0.117
& 1.092\(\pm\)0.059
& 1.178\(\pm\)0.100
& 1.094\(\pm\)0.037
& 1.093\(\pm\)0.050
& 1.122\(\pm\)0.009\\
& \(P_{\mathrm{CON}}\)
& 1.013\(\pm\)0.056
& 1.019\(\pm\)0.074
& 1.069\(\pm\)0.066
& 1.027\(\pm\)0.044
& 1.106\(\pm\)0.098
& 1.014\(\pm\)0.044
& 1.043\(\pm\)0.050
& \textbf{1.000\(\pm\)0.036}
& 1.030\(\pm\)0.051
& 1.373\(\pm\)0.097\\
& NFE
& \textbf{9251\(\pm\)189}
& 11616\(\pm\)557
& 11088\(\pm\)850
& 11620\(\pm\)516
& 11044\(\pm\)815
& 11396\(\pm\)325
& 11308\(\pm\)510
& 11352\(\pm\)347
& 11396\(\pm\)385
& 11161\(\pm\)65\\
\midrule

\multirow{3}{*}{\(G_5\)}
& \(E_{\mathrm{MCE}}\)
& 1.138\(\pm\)0.201
& 1.022\(\pm\)0.038
& 1.039\(\pm\)0.032
& \textbf{1.000\(\pm\)0.064}
& 1.046\(\pm\)0.053
& 1.050\(\pm\)0.157
& 1.033\(\pm\)0.049
& 1.055\(\pm\)0.049
& 1.069\(\pm\)0.084
& 1.072\(\pm\)0.001\\
& \(P_{\mathrm{CON}}\)
& 1.117\(\pm\)0.093
& \textbf{1.000\(\pm\)0.049}
& 1.088\(\pm\)0.053
& 1.030\(\pm\)0.070
& 1.100\(\pm\)0.048
& 1.029\(\pm\)0.041
& 1.064\(\pm\)0.052
& 1.046\(\pm\)0.037
& 1.078\(\pm\)0.044
& 1.234\(\pm\)0.065\\
& NFE
& \textbf{9011\(\pm\)96}
& 10912\(\pm\)454
& 10824\(\pm\)516
& 10781\(\pm\)475
& 11088\(\pm\)875
& 10736\(\pm\)473
& 11176\(\pm\)629
& 10560\(\pm\)622
& 11176\(\pm\)694
& 11162\(\pm\)62\\
\midrule

\multirow{3}{*}{\(G_6\)}
& \(E_{\mathrm{MCE}}\)
& \textbf{1.000\(\pm\)0.037}
& 1.021\(\pm\)0.022
& 1.047\(\pm\)0.089
& 1.030\(\pm\)0.019
& 1.080\(\pm\)0.074
& 1.081\(\pm\)0.145
& 1.102\(\pm\)0.163
& 1.032\(\pm\)0.029
& 1.112\(\pm\)0.173
& 1.067\(\pm\)0.024\\
& \(P_{\mathrm{CON}}\)
& 1.044\(\pm\)0.042
& 1.010\(\pm\)0.053
& 1.019\(\pm\)0.075
& 1.004\(\pm\)0.037
& 1.066\(\pm\)0.066
& 1.010\(\pm\)0.073
& 1.006\(\pm\)0.039
& \textbf{1.000\(\pm\)0.040}
& 1.018\(\pm\)0.064
& 1.223\(\pm\)0.131\\
& NFE
& \textbf{8989\(\pm\)151}
& 10516\(\pm\)527
& 11176\(\pm\)473
& 10736\(\pm\)594
& 10736\(\pm\)594
& 10560\(\pm\)549
& 11352\(\pm\)615
& 10780\(\pm\)631
& 11088\(\pm\)742
& 11160\(\pm\)54\\
\midrule

\multirow{3}{*}{\(G_7\)}
& \(E_{\mathrm{MCE}}\)
& \textbf{1.000\(\pm\)0.054}
& 1.027\(\pm\)0.080
& 1.070\(\pm\)0.058
& 1.007\(\pm\)0.058
& 1.118\(\pm\)0.126
& 1.052\(\pm\)0.054
& 1.099\(\pm\)0.010
& 1.094\(\pm\)0.047
& 1.091\(\pm\)0.023
& 1.107\(\pm\)0.001\\
& \(P_{\mathrm{CON}}\)
& 1.016\(\pm\)0.034
& \textbf{1.000\(\pm\)0.031}
& 1.044\(\pm\)0.045
& 1.015\(\pm\)0.023
& 1.035\(\pm\)0.041
& 1.015\(\pm\)0.048
& 1.031\(\pm\)0.018
& 1.038\(\pm\)0.028
& 1.040\(\pm\)0.026
& 1.069\(\pm\)0.059\\
& NFE
& \textbf{9181\(\pm\)85}
& 10560\(\pm\)359
& 10604\(\pm\)385
& 11000\(\pm\)293
& 10868\(\pm\)510
& 10692\(\pm\)510
& 11132\(\pm\)856
& 10604\(\pm\)325
& 11176\(\pm\)629
& 11177\(\pm\)42\\
\midrule

\multirow{3}{*}{\(G_8\)}
& \(E_{\mathrm{MCE}}\)
& \textbf{1.000\(\pm\)0.052}
& 1.063\(\pm\)0.010
& 1.085\(\pm\)0.111
& 1.057\(\pm\)0.029
& 1.077\(\pm\)0.026
& 1.092\(\pm\)0.078
& 1.061\(\pm\)0.034
& 1.063\(\pm\)0.014
& 1.065\(\pm\)0.016
& 1.094\(\pm\)0.086\\
& \(P_{\mathrm{CON}}\)
& 1.040\(\pm\)0.031
& 1.022\(\pm\)0.030
& 1.029\(\pm\)0.041
& 1.017\(\pm\)0.036
& \textbf{1.000\(\pm\)0.053}
& 1.020\(\pm\)0.029
& 1.018\(\pm\)0.028
& 1.014\(\pm\)0.048
& 1.051\(\pm\)0.054
& 1.142\(\pm\)0.130\\
& NFE
& \textbf{9149\(\pm\)104}
& 11264\(\pm\)308
& 11352\(\pm\)742
& 10912\(\pm\)455
& 11132\(\pm\)624
& 11308\(\pm\)510
& 11000\(\pm\)464
& 11220\(\pm\)374
& 11044\(\pm\)671
& 11192\(\pm\)48\\
\midrule

\multirow{3}{*}{\(G_9\)}
& \(E_{\mathrm{MCE}}\)
& 1.012\(\pm\)0.058
& 1.061\(\pm\)0.153
& 1.061\(\pm\)0.103
& \textbf{1.000\(\pm\)0.039}
& 1.047\(\pm\)0.059
& 1.031\(\pm\)0.016
& 1.049\(\pm\)0.107
& 1.046\(\pm\)0.119
& 1.074\(\pm\)0.103
& 1.126\(\pm\)0.143\\
& \(P_{\mathrm{CON}}\)
& 1.015\(\pm\)0.048
& 1.021\(\pm\)0.049
& 1.013\(\pm\)0.038
& 1.009\(\pm\)0.046
& 1.033\(\pm\)0.040
& \textbf{1.000\(\pm\)0.035}
& 1.005\(\pm\)0.038
& 1.015\(\pm\)0.030
& 1.048\(\pm\)0.075
& 1.276\(\pm\)0.186\\
& NFE
& \textbf{9081\(\pm\)142}
& 10868\(\pm\)466
& 11528\(\pm\)1011
& 10780\(\pm\)475
& 10877\(\pm\)654
& 10780\(\pm\)596
& 11264\(\pm\)809
& 10736\(\pm\)516
& 11133\(\pm\)830
& 11165\(\pm\)51\\

\midrule
\multicolumn{12}{l}{
\textit{Statistical summary based on \(E_{\mathrm{MCE}}\)}}\\
\multicolumn{2}{l}{Avg. rank\(\downarrow\)}
& \textbf{2.222}
& 4.222
& 6.056
& 2.333
& 7.389
& 5.722
& 6.778
& 5.000
& 7.056
& 8.222\\
\multicolumn{2}{l}{Sign-test \(p\) vs PosEDO}
& --
& 0.039
& 0.180
& 0.180
& 0.039
& 0.039
& 0.039
& 0.039
& 0.039
& 0.039\\
\multicolumn{2}{l}{W/T/L vs PosEDO}
& --
& 8/0/1
& 7/0/2
& 7/0/2
& 8/0/1
& 8/0/1
& 8/0/1
& 8/0/1
& 8/0/1
& 8/0/1\\

\midrule
\multicolumn{12}{l}{
\textit{Statistical summary based on \(P_{\mathrm{CON}}\)}}\\
\multicolumn{2}{l}{Avg. rank\(\downarrow\)}
& 6.389
& 3.611
& 6.889
& 3.167
& 6.611
& \textbf{2.944}
& 5.333
& 3.167
& 6.889
& 10.000\\
\multicolumn{2}{l}{Sign-test \(p\) vs PosEDO}
& --
& 0.180
& 0.508
& 0.039
& 0.508
& 0.039
& 1.000
& 0.070
& 1.000
& 0.004\\
\multicolumn{2}{l}{W/T/L vs PosEDO}
& --
& 2/0/7
& 3/0/6
& 1/0/8
& 6/0/3
& 1/0/8
& 4/0/5
& 1/1/7
& 5/0/4
& 9/0/0\\

\midrule
\multicolumn{12}{l}{
\textit{Statistical summary based on NFE}}\\
\multicolumn{2}{l}{Avg. rank\(\downarrow\)}
& \textbf{1.000}
& 4.667
& 6.889
& 5.056
& 5.500
& 4.611
& 7.889
& 4.389
& 7.444
& 7.556\\
\multicolumn{2}{l}{Sign-test \(p\) vs PosEDO}
& --
& 0.004
& 0.004
& 0.004
& 0.004
& 0.004
& 0.004
& 0.004
& 0.004
& 0.004\\
\multicolumn{2}{l}{W/T/L vs PosEDO}
& --
& 9/0/0
& 9/0/0
& 9/0/0
& 9/0/0
& 9/0/0
& 9/0/0
& 9/0/0
& 9/0/0
& 9/0/0\\
\bottomrule
\end{tabular}
\end{adjustbox}
}
\end{table*}

\begin{table}[htbp]
\centering
\caption{Overall change-detection performance on the PGPS-6D benchmark. Precision, recall, F1, and the average number of detected change points (Detected) are averaged over the nine instances and 10 independent runs per instance. A detected change point is counted as correct only when it exactly matches a ground-truth change point, corresponding to a tolerance window of zero. Higher precision, recall, and F1 values are better. Each instance contains five ground-truth change points, and hence a Detected value closer to \(5\) is preferred. DBD-Rand results are reported for all tested \(\alpha\) values.}
\label{tab:pgps-change-detection-overall-all-alpha}
{\scriptsize
\begin{adjustbox}{max width=0.5\textwidth, keepaspectratio}
\begin{tabular}{lcccc}
\toprule
\textbf{Method} & \textbf{Precision} & \textbf{Recall} & \textbf{F1} & \textbf{Detected}\\
\midrule
PosEDO & \textbf{0.822$\pm$0.260} & 0.816$\pm$0.261 & \textbf{0.819$\pm$0.260} & \textbf{4.933$\pm$0.100}\\
PosEDO-CD & 0.809$\pm$0.223 & 0.791$\pm$0.233 & 0.799$\pm$0.228 & 4.856$\pm$0.159\\
FBCD-Arch & 0.578$\pm$0.099 & 0.753$\pm$0.135 & 0.646$\pm$0.105 & 6.689$\pm$0.855\\
MFBCD-Arch & 0.402$\pm$0.064 & 0.573$\pm$0.097 & 0.467$\pm$0.075 & 7.222$\pm$0.650\\
FBCD-DETOInit & 0.564$\pm$0.095 & 0.751$\pm$0.129 & 0.638$\pm$0.100 & 6.800$\pm$0.711\\
MFBCD-DETOInit & 0.376$\pm$0.079 & 0.513$\pm$0.112 & 0.428$\pm$0.091 & 6.922$\pm$0.435\\
FBCD-NNIT & 0.576$\pm$0.125 & 0.740$\pm$0.145 & 0.641$\pm$0.127 & 6.589$\pm$0.788\\
MFBCD-NNIT & 0.418$\pm$0.077 & 0.613$\pm$0.099 & 0.493$\pm$0.086 & 7.478$\pm$0.295\\
FBCD-Rand & 0.550$\pm$0.109 & 0.716$\pm$0.157 & 0.616$\pm$0.125 & 6.633$\pm$0.652\\
MFBCD-Rand & 0.413$\pm$0.069 & 0.600$\pm$0.087 & 0.485$\pm$0.075 & 7.367$\pm$0.458\\
DBD-Rand($\alpha=0$) & 0.294$\pm$0.000 & \textbf{1.000$\pm$0.000} & 0.455$\pm$0.000 & 17.000$\pm$0.000\\
DBD-Rand($\alpha=0.2$) & 0.257$\pm$0.077 & 0.680$\pm$0.150 & 0.346$\pm$0.073 & 11.500$\pm$2.445\\
DBD-Rand($\alpha=0.4$) & 0.140$\pm$0.198 & 0.109$\pm$0.123 & 0.088$\pm$0.114 & 1.456$\pm$1.308\\
DBD-Rand($\alpha=0.6$) & 0.109$\pm$0.159 & 0.093$\pm$0.107 & 0.072$\pm$0.095 & 1.200$\pm$1.130\\
DBD-Rand($\alpha=0.8$) & 0.069$\pm$0.097 & 0.064$\pm$0.080 & 0.046$\pm$0.062 & 0.911$\pm$1.025\\
DBD-Rand($\alpha=1.0$) & 0.035$\pm$0.074 & 0.058$\pm$0.078 & 0.035$\pm$0.058 & 0.844$\pm$0.994\\
\bottomrule
\end{tabular}
\end{adjustbox}
}
\end{table}

\subsection{Performance in Higher-dimensional Parameter Spaces}

Fig.~\ref{fig:pgps-high-dimensional} evaluates the methods as the
parameter dimension increases from \(12\) to \(48\). PosEDO achieves the
best average \(E_{\mathrm{MCE}}\) rank under all four dimensional
settings, indicating that its calibration advantage remains stable as
the parameter space expands.

For change detection, PosEDO obtains the highest F1 score from \(12\)D
to \(36\)D. At \(48\)D, its F1 remains competitive, although several
baselines achieve slightly higher values. Overall, these results show
that PosEDO remains effective in higher-dimensional parameter spaces,
while the change-detection advantage gradually narrows as dimensionality
increases.

\section{Conclusions and Future Work}

This paper formulated online regime-aware calibration of black-box
social simulators as an observation-driven DOP, in which the objective
is determined by sequential observations and the detected calibration
window. In this setting, ordinary observation updates can alter fitness
values even without a regime change, while the relationship between
successive regimes is generally unknown. These properties make
conventional fitness-based detection and history-based population
adaptation unreliable or costly.

To address these challenges, we proposed PosEDO, which learns an
observation-conditioned distribution over simulator parameters from the parameter--trajectory pairs generated during evolutionary search.
Posterior shifts provide a fitness-independent signal for change detection, while posterior samples provide candidate parameters for population adaptation. 
The same parameter–trajectory records are reused for online
posterior training, so posterior learning, detection, and adaptation
introduce no additional simulator calls. Experiments on nonstationary
economic and financial simulators show that PosEDO improves calibration
accuracy and change-detection quality while requiring fewer simulator
evaluations. It also achieves strong convergence performance on the BH
benchmark and maintains effective calibration and detection performance
as the PGPS parameter dimension increases.

The current study uses fixed observation lengths and empirically selected detection thresholds. Its posterior estimator also becomes more difficult
to learn as simulator stochasticity and parameter dimensionality increase.
Future work will investigate adaptive observation windows and detection
thresholds, more robust and more simulation-efficient posterior learning, and
extensions to constrained, multi-objective, and multi-fidelity
observation-driven DOPs.

\bibliographystyle{IEEEtran}
\bibliography{references}

@article{ref1,
  title={The economy needs agent-based modelling},
  author={Farmer, J Doyne and Foley, Duncan},
  journal={Nature},
  volume={460},
  number={7256},
  pages={685--686},
  year={2009},
  publisher={Nature Publishing Group UK London}
}

@article{ref2,
  title={Heterogeneous beliefs and routes to chaos in a simple asset pricing model},
  author={Brock, William A and Hommes, Cars H},
  journal={Journal of Economic dynamics and Control},
  volume={22},
  number={8-9},
  pages={1235--1274},
  year={1998},
  publisher={Elsevier}
}

@article{ref5,
  title={Strategies for containing an emerging influenza pandemic in Southeast Asia},
  author={Ferguson, Neil M and Cummings, Derek AT and Cauchemez, Simon and Fraser, Christophe and Riley, Steven and Meeyai, Aronrag and Iamsirithaworn, Sopon and Burke, Donald S},
  journal={Nature},
  volume={437},
  number={7056},
  pages={209--214},
  year={2005},
  publisher={Nature Publishing Group UK London}
}

@article{ref8,
  title={Estimation of ergodic agent-based models by simulated minimum distance},
  author={Grazzini, Jakob and Richiardi, Matteo},
  journal={Journal of Economic Dynamics and Control},
  volume={51},
  pages={148--165},
  year={2015},
  publisher={Elsevier}
}

@article{ref9,
  title={A comparison of economic agent-based model calibration methods},
  author={Platt, Donovan},
  journal={Journal of Economic Dynamics and Control},
  volume={113},
  pages={103859},
  year={2020},
  publisher={Elsevier}
}

@inproceedings{altamirano2023robust,
  title={Robust and scalable Bayesian online changepoint detection},
  author={Altamirano, Matias and Briol, Fran{\c{c}}ois-Xavier and Knoblauch, Jeremias},
  booktitle={International Conference on Machine Learning},
  pages={642--663},
  year={2023},
  organization={PMLR}
}

@ARTICLE{8246564,
  author={Wang, Shuo and Minku, Leandro L. and Yao, Xin},
  journal={IEEE Transactions on Neural Networks and Learning Systems}, 
  title={A Systematic Study of Online Class Imbalance Learning With Concept Drift}, 
  year={2018},
  volume={29},
  number={10},
  pages={4802-4821}
}

@ARTICLE{6148271,
  author={Nguyen, Trung Thanh and Yao, Xin},
  journal={IEEE Transactions on Evolutionary Computation}, 
  title={Continuous Dynamic Constrained Optimization—The Challenges}, 
  year={2012},
  volume={16},
  number={6},
  pages={769-786}
}

@inproceedings{alami2023bayesian,
  title={Bayesian change-point detection for bandit feedback in non-stationary environments},
  author={Alami, Reda},
  booktitle={Asian Conference on Machine Learning},
  pages={17--31},
  year={2023},
  organization={PMLR}
}

@article{ref13,
  title={Automatic calibration of dynamic and heterogeneous parameters in agent-based models},
  author={Kim, Dongjun and Yun, Tae-Sub and Moon, Il-Chul and Bae, Jang Won},
  journal={Autonomous Agents and Multi-Agent Systems},
  volume={35},
  number={2},
  pages={46},
  year={2021},
  publisher={Springer}
}

@article{ref15,
  title={Data assimilation in the geosciences: An overview of methods, issues, and perspectives},
  author={Carrassi, Alberto and Bocquet, Marc and Bertino, Laurent and Evensen, Geir},
  journal={Wiley Interdisciplinary Reviews: Climate Change},
  volume={9},
  number={5},
  pages={e535},
  year={2018},
  publisher={Wiley Online Library}
}

@article{ref16,
  title={Evolutionary dynamic optimization: A survey of the state of the art},
  author={Nguyen, Trung Thanh and Yang, Shengxiang and Branke, Juergen},
  journal={Swarm and Evolutionary Computation},
  volume={6},
  pages={1--24},
  year={2012},
  publisher={Elsevier}
}

@article{ref18,
  title={A survey of evolutionary continuous dynamic optimization over two decades—Part A},
  author={Yazdani, Danial and Cheng, Ran and Yazdani, Donya and Branke, J{\"u}rgen and Jin, Yaochu and Yao, Xin},
  journal={IEEE Transactions on Evolutionary Computation},
  volume={25},
  number={4},
  pages={609--629},
  year={2021},
  publisher={IEEE}
}

@inproceedings{ref20,
  title={Memory enhanced evolutionary algorithms for changing optimization problems},
  author={Branke, J{\"u}rgen},
  booktitle={Proceedings of the 1999 Congress on Evolutionary Computation-CEC99 (Cat. No. 99TH8406)},
  volume={3},
  pages={1875--1882},
  year={1999},
  organization={IEEE}
}

@inproceedings{ref21,
  title={Adaptive particle swarm optimization: detection and response to dynamic systems},
  author={Hu, Xiaohui and Eberhart, Russell C},
  booktitle={Proceedings of the 2002 congress on evolutionary computation. CEC'02 (cat. No. 02TH8600)},
  volume={2},
  pages={1666--1670},
  year={2002},
  organization={IEEE}
}

@article{ref22,
  title={Population-based incremental learning with associative memory for dynamic environments},
  author={Yang, Shengxiang and Yao, Xin},
  journal={IEEE Transactions on evolutionary computation},
  volume={12},
  number={5},
  pages={542--561},
  year={2008},
  publisher={IEEE}
}

@article{ref23,
  title={Neural network-based information transfer for dynamic optimization},
  author={Liu, Xiao-Fang and Zhan, Zhi-Hui and Gu, Tian-Long and Kwong, Sam and Lu, Zhenyu and Duh, Henry Been-Lirn and Zhang, Jun},
  journal={IEEE transactions on neural networks and learning systems},
  volume={31},
  number={5},
  pages={1557--1570},
  year={2019},
  publisher={IEEE}
}

@article{ref24,
  title={Neural network for change direction prediction in dynamic optimization},
  author={Liu, Xiao-Fang and Zhan, Zhi-Hui and Zhang, Jun},
  journal={IEEE access},
  volume={6},
  pages={72649--72662},
  year={2018},
  publisher={IEEE}
}

@article{ref25,
  title={Kalman-extended genetic algorithm for search in nonstationary environments with noisy fitness evaluations},
  author={Stroud, Phillip D.},
  journal={IEEE Transactions on Evolutionary Computation},
  volume={5},
  number={1},
  pages={66--77},
  year={2002},
  publisher={IEEE}
}

@article{ref25-1,
  title={A new prediction strategy for dynamic multi-objective optimization using diffusion model},
  author={Wang, Feng and Xie, Jinsong and Zhou, Aimin and Tang, Ke},
  journal={IEEE Transactions on Evolutionary Computation},
  year={2025},
  publisher={IEEE}
}

@article{ref25-3,
  title={A data-driven evolutionary transfer optimization for expensive problems in dynamic environments},
  author={Li, Ke and Chen, Renzhi and Yao, Xin},
  journal={IEEE Transactions on Evolutionary Computation},
  year={2023},
  publisher={IEEE}
}

@article{ref27,
  title={An adaptive multipopulation framework for locating and tracking multiple optima},
  author={Li, Changhe and Nguyen, Trung Thanh and Yang, Ming and Mavrovouniotis, Michalis and Yang, Shengxiang},
  journal={IEEE transactions on evolutionary computation},
  volume={20},
  number={4},
  pages={590--605},
  year={2015},
  publisher={IEEE}
}

@article{refyx2024,
  title={Engine calibration with surrogate-assisted bilevel evolutionary algorithm},
  author={Yu, Xunzhao and Wang, Yan and Zhu, Ling and Filev, Dimitar and Yao, Xin},
  journal={IEEE Transactions on Cybernetics},
  volume={54},
  number={6},
  pages={3832--3845},
  year={2023},
  publisher={IEEE}
}

@article{ref30,
  title={Fast $\varepsilon$-free inference of simulation models with bayesian conditional density estimation},
  author={Papamakarios, George and Murray, Iain},
  journal={Advances in neural information processing systems},
  volume={29},
  year={2016}
}

@inproceedings{refyx2001,
  title={Evolutionary design calibration},
  author={Schnier, Thorsten and Yao, Xin},
  booktitle={International Conference on Evolvable Systems},
  pages={26--37},
  year={2001},
  organization={Springer}
}

@article{ref33,
  title={Approximate Bayesian computation in population genetics},
  author={Beaumont, Mark A and Zhang, Wenyang and Balding, David J},
  journal={Genetics},
  volume={162},
  number={4},
  pages={2025--2035},
  year={2002},
  publisher={Oxford University Press}
}

@inproceedings{ref35,
  title={Automatic posterior transformation for likelihood-free inference},
  author={Greenberg, David and Nonnenmacher, Marcel and Macke, Jakob},
  booktitle={International conference on machine learning},
  pages={2404--2414},
  year={2019},
  organization={PMLR}
}

@article{ref-rjj,
  title={Towards calibrating financial market simulators with high-frequency data},
  author={Yang, Peng and Ren, Junji and Wang, Feng and Tang, Ke},
  journal={Complex System Modeling and Simulation},
  year={2025},
  publisher={TUP}
}

@article{ref-zjj,
  author={Zhao, Junjie and Zhang, Chengxi and Qin, Min and Yang, Peng},
  journal={IEEE Transactions on Signal Processing}, 
  title={QuantFactor REINFORCE: Mining Steady Formulaic Alpha Factors With Variance-Bounded REINFORCE}, 
  year={2025},
  volume={73},
  number={},
  pages={2448-2463}
}

@article{ref36,
  title={A method of simulated moments for estimation of discrete response models without numerical integration},
  author={McFadden, Daniel},
  journal={Econometrica: Journal of the Econometric Society},
  pages={995--1026},
  year={1989},
  publisher={JSTOR}
}

@inproceedings{ref38,
  title={Efficient calibration of multi-agent simulation models from output series with bayesian optimization},
  author={Bai, Yuanlu and Lam, Henry and Balch, Tucker and Vyetrenko, Svitlana},
  booktitle={Proceedings of the Third ACM International Conference on AI in Finance},
  pages={437--445},
  year={2022}
}

@article{ref39,
  author={Wang, Chenkai and Ren, Junji and Yang, Peng},
  journal={IEEE Transactions on Computational Social Systems}, 
  title={Alleviating Nonidentifiability: A High-Fidelity Calibration Objective for Financial Market Simulation With Multivariate Time Series Data}, 
  year={2025},
  volume={12},
  number={6},
  pages={4910-4922}
}

@article{ref41,
  title={Dynamic toll pricing using dynamic traffic assignment system with online calibration},
  author={Zhang, Yundi and Atasoy, Bilge and Akkinepally, Arun and Ben-Akiva, Moshe},
  journal={Transportation Research Record},
  volume={2673},
  number={10},
  pages={532--546},
  year={2019},
  publisher={SAGE Publications Sage CA: Los Angeles, CA}
}

@inproceedings{ref42,
  title={Tracking changing extrema with adaptive particle swarm optimizer},
  author={Carlisle, Anthony and Dozler, G},
  booktitle={Proceedings of the 5th biannual world automation congress},
  volume={13},
  pages={265--270},
  year={2002},
  organization={IEEE}
}

@article{ref44,
  title={Dynamic evolutionary algorithm with variable relocation},
  author={Woldesenbet, Yonas G and Yen, Gary G},
  journal={IEEE Transactions on Evolutionary Computation},
  volume={13},
  number={3},
  pages={500--513},
  year={2009},
  publisher={IEEE}
}

@article{ref45,
  title={Differential evolution for dynamic environments with unknown numbers of optima},
  author={Du Plessis, Mathys C and Engelbrecht, Andries P},
  journal={Journal of Global Optimization},
  volume={55},
  number={1},
  pages={73--99},
  year={2013},
  publisher={Springer}
}

@article{ref47,
  title={Multi-population methods in unconstrained continuous dynamic environments: The challenges},
  author={Li, Changhe and Nguyen, Trung Thanh and Yang, Ming and Yang, Shengxiang and Zeng, Sanyou},
  journal={Information Sciences},
  volume={296},
  pages={95--118},
  year={2015},
  publisher={Elsevier}
}

@article{ref48,
  title={Multi-agent-based order book model of financial markets},
  author={Preis, Tobias and Golke, Sebastian and Paul, Wolfgang and Schneider, Johannes J},
  journal={Europhysics Letters},
  volume={75},
  number={3},
  pages={510},
  year={2006},
  publisher={IOP Publishing}
}

@article{yu2025review,
  title={Review on system identification, control, and optimization based on artificial intelligence},
  author={Yu, Pan and Wan, Hui and Zhang, Bozhi and Wu, Qiang and Zhao, Bohao and Xu, Chen and Yang, Shangbin},
  journal={Mathematics},
  volume={13},
  number={6},
  pages={952},
  year={2025},
  publisher={MDPI}
}

@book{billings2013nonlinear,
  title={Nonlinear system identification: NARMAX methods in the time, frequency, and spatio-temporal domains},
  author={Billings, Stephen A},
  year={2013},
  publisher={John Wiley \& Sons}
}

@article{he2021,
  title={A Kalman particle filter for online parameter estimation with applications to affine models},
  author={He, Jian and Khedher, Asma and Spreij, Peter},
  journal={Statistical Inference for Stochastic Processes},
  volume={24},
  number={2},
  pages={353--403},
  year={2021},
  publisher={Springer}
}

@article{lu2018learning,
  title={Learning under concept drift: A review},
  author={Lu, Jie and Liu, Anjin and Dong, Fan and Gu, Feng and Gama, Joao and Zhang, Guangquan},
  journal={IEEE transactions on knowledge and data engineering},
  volume={31},
  number={12},
  pages={2346--2363},
  year={2018},
  publisher={IEEE}
}

@article{hinder2024oneA,
  title={One or two things we know about concept drift—a survey on monitoring in evolving environments. Part A: detecting concept drift},
  author={Hinder, Fabian and Vaquet, Valerie and Hammer, Barbara},
  journal={Frontiers in Artificial Intelligence},
  volume={7},
  pages={1330257},
  year={2024},
  publisher={Frontiers Media SA}
}

@article{hinder2024oneB,
  title={One or two things we know about concept drift—a survey on monitoring in evolving environments. Part B: locating and explaining concept drift},
  author={Hinder, Fabian and Vaquet, Valerie and Hammer, Barbara},
  journal={Frontiers in Artificial Intelligence},
  volume={7},
  pages={1330258},
  year={2024},
  publisher={Frontiers Media SA}
}

@article{papamakarios2017masked,
  title={Masked autoregressive flow for density estimation},
  author={Papamakarios, George and Pavlakou, Theo and Murray, Iain},
  journal={Advances in neural information processing systems},
  volume={30},
  year={2017}
}

@article{greco2025unsupervised,
  title={Unsupervised concept drift detection from deep learning representations in real-time},
  author={Greco, Salvatore and Vacchetti, Bartolomeo and Apiletti, Daniele and Cerquitelli, Tania},
  journal={IEEE Transactions on Knowledge and Data Engineering},
  year={2025},
  publisher={IEEE}
}

@article{ruan2024knowledge,
  title={Knowledge transfer for dynamic multi-objective optimization with a changing number of objectives},
  author={Ruan, Gan and Minku, Leandro L and Menzel, Stefan and Sendhoff, Bernhard and Yao, Xin},
  journal={IEEE Transactions on Emerging Topics in Computational Intelligence},
  volume={8},
  number={6},
  pages={4210--4224},
  year={2024},
  publisher={IEEE}
}

@article{ruan2024learning,
  title={Learning to expand/contract pareto sets in dynamic multi-objective optimization with a changing number of objectives},
  author={Ruan, Gan and Minku, Leandro L and Menzel, Stefan and Sendhoff, Bernhard and Yao, Xin},
  journal={IEEE Transactions on Evolutionary Computation},
  year={2024},
  publisher={IEEE}
}

@article{yang2024reducing,
  title={Reducing idleness in financial cloud services via multi-objective evolutionary reinforcement learning based load balancer},
  author={Yang, Peng and Zhang, Laoming and Liu, Haifeng and Li, Guiying},
  journal={Science China Information Sciences},
  volume={67},
  number={2},
  pages={120102},
  year={2024},
  publisher={Springer}
}

@article{nasiri2016history,
  title={History-driven firefly algorithm for optimisation in dynamic and uncertain environments},
  author={Nasiri, Babak and Meybodi, Mohammad Reza},
  journal={International Journal of Bio-Inspired Computation},
  volume={8},
  number={5},
  pages={326--339},
  year={2016},
  publisher={Inderscience Publishers (IEL)}
}

@inproceedings{niu2019bacterial,
  title={Bacterial foraging optimization with memory and clone schemes for dynamic environments},
  author={Niu, Ben and Liu, Qianying and Wang, Jun},
  booktitle={International conference on swarm intelligence},
  pages={352--360},
  year={2019},
  organization={Springer}
}

\newpage

\vfill

\end{document}